%% file: main.tex
\documentclass[10pt,twocolumn,letterpaper]{article}

\usepackage{cvpr}              
\usepackage{amsmath}
\usepackage{amssymb}
\usepackage{comment}
\usepackage{rotating}
\usepackage{adjustbox}
\usepackage{svg}
\usepackage{array}
\usepackage{float}
\usepackage{graphicx}
\usepackage{import}
\usepackage{xcolor}

\definecolor{bestblue}{HTML}{3a7bc1}
\definecolor{bestgreen}{HTML}{41914f}

\definecolor{cvprblue}{rgb}{0.21,0.49,0.74}
\usepackage[pagebackref,breaklinks,colorlinks,citecolor=cvprblue]{hyperref}

\def\paperID{12185} 
\def\confName{CVPR}
\def\confYear{2024}

\title{RNb-NeuS: Reflectance and Normal-based Multi-View 3D Reconstruction}

\author{Baptiste Brument$^{1,}$\thanks{Equal contributions. {\tt\scriptsize brument.bcb@gmail.com} / {\tt\scriptsize rb@di.ku.dk}}\footnotemark[1]
\and
Robin Bruneau$^{1,2,}$\footnotemark[1]
\and
Yvain Quéau$^3$
\and
Jean Mélou$^1$
\and
François Bernard Lauze$^2$
\and
Jean-Denis Durou$^1$
\and
Lilian Calvet$^4$
\and
  \\$^1$IRIT, UMR CNRS 5505, Toulouse, France\\
  $^2$DIKU, Copenhagen, Denmark\\
  $^3$Normandie Univ, UNICAEN, ENSICAEN, CNRS, GREYC, Caen, France\\
  $^4$OR-X, Balgrist Hospital, University of Zurich, Zürich, Switzerland
}

\begin{document}
\maketitle
\input{sec/0_abstract}    
\input{sec/1_intro}

\input{sec/2_sota}

\input{sec/3_method}
\input{sec/4_experiments}
\input{sec/5_conclusion}

\input{sec/X_suppl}

{
    \small
    \bibliographystyle{ieeenat_fullname}
    \bibliography{biblio_propre}
}


\end{document}

%% file: sec/0_abstract.tex
\begin{abstract}

This paper introduces a versatile paradigm for integrating multi-view reflectance (optional) and normal maps acquired through photometric stereo. Our approach employs a pixel-wise joint re-parameterization of reflectance and normal, considering them as a vector of radiances rendered under simulated, varying illumination. This re-parameterization enables the seamless integration of reflectance and normal maps as input data in neural volume rendering-based 3D reconstruction while preserving a single optimization objective. In contrast,  recent multi-view photometric stereo (MVPS) methods depend on multiple, potentially conflicting objectives. Despite its apparent simplicity, our proposed approach outperforms state-of-the-art approaches in MVPS benchmarks across F-score, Chamfer distance, and mean angular error metrics. Notably, it significantly improves the detailed 3D reconstruction of areas with high curvature or low visibility.
\end{abstract}

%% file: sec/1_intro.tex
\section{Introduction}
\label{sec:intro}

Automatic 3D reconstruction is pivotal in various fields, such as archaeological and cultural heritage (virtual reconstruction), medical imaging (surgical planning), virtual and augmented reality, games and film production.

Multi-view stereo (MVS)~\cite{FurukawaP15}, which retrieves the geometry of a scene seen from multiple viewpoints, is the most famous 3D reconstruction solution. Coupled with neural volumetric rendering (NVR) techniques~\cite{WangLLTKW21}, it effectively handles complex structures and self-occlusions. However, dealing with non-Lambertian scenes remains a challenge due to the breakdown of the underlying brightness consistency assumption. The problem is also ill-posed in certain configurations e.g., poorly textured scene~\cite{XuKTP23} or degenerate viewpoints configurations with limited baselines. Moreover, despite recent efforts in this direction~\cite{li2023neuralangelo}, recovering the thinnest geometric details remains difficult under fixed illumination. In such a setting, estimating the reflectance of the scene also remains a challenge. 

\begin{figure}[t]
    \includegraphics[width=\linewidth]{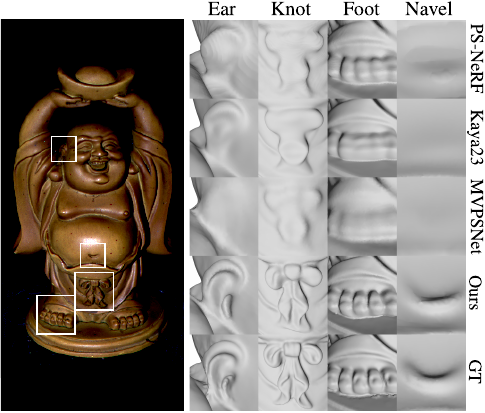}
    \caption{One image from DiLiGenT-MV's Buddha dataset~\cite{LiZWSDT20}, and 3D reconstruction results from several recent MVPS methods: \cite{YangCCCW22,KayaKOFG23,mvpsnet} and ours. The latter provides the fine details closest to the ground truth (GT), while being remarkably simpler.
    }
    \label{fig:example}
\end{figure}

On the other hand, photometric stereo (PS)~\cite{Woodham79}, which relies on a collection of images acquired under varying lighting, excels in the recovery of high-frequency details under the form of normal maps. It is also the only photographic technique that can estimate reflectance. And, with the recent advent of deep learning techniques~\cite{sdmunips}, PS gained enough maturity to handle non-Lambertian surfaces and complex illumination. 
Yet, its reconstruction of geometry's low frequencies remains suboptimal.

Given these complementary characteristics, the integration of MVS and PS seems natural. This integration, known as multi-view photometric stereo (MVPS), aims to reconstruct geometry from multiple views and illumination conditions. Recent MVPS solutions jointly solve MVS and PS within a multi-objective optimization, potentially losing the thinnest details due to the possible incompatibility of these objectives -- see Fig.~\ref{fig:example}. In this work, we explore a simpler route for solving MVPS by decoupling the two problems. 

We start with the observation that recent PS techniques deliver exceptionally high-quality reflectance and normal maps, which we use as input data. To accurately reconstruct the surface reflectance and geometry, we need to fuse these maps, a challenging task within a single-objective optimization due to their inhomogeneity.  Our method provides a solution to this problem by combining NVR with a simple and effective pixel-wise re-parameterization.

In this method, the input reflectance and normal for each pixel are merged into a vector of radiances simulated under arbitrary, varying illumination. We then adapt an NVR pipeline to optimize the consistency of these simulations wrt to the scene reflectance and geometry, modeled as the zero-level set of a trained signed distance function (SDF). Coupled with a state-of-the-art PS method such as~\cite{sdmunips} for obtaining the input reflectance and normals, this approach yields an MVPS pipeline reaching an unprecedented level of fine details, as illustrated in Fig.~\ref{fig:example}. Besides being the first to exploit reflectance as a prior, our proposed MVPS paradigm is extremely versatile, 
compatible with any existing or future PS method, whether calibrated or uncalibrated, deep learning-based, or classic optimization procedures.

The rest of this work is organized as follows. Sect.~\ref{sec:sota} discusses state-of-the-art MVPS methods. The proposed 3D reconstruction from reflectance and normals is detailed in Sect.~\ref{sec:method}. Sect.~\ref{sec:mvps} then sketches a proposal for an MVPS algorithm based on this approach. Sect.~\ref{sec:xp} extensively evaluates this algorithm, before our conclusions are drawn in Sect.~\ref{sec:concl}.


%% file: sec/2_sota.tex
\section{Related work}
\label{sec:sota}

\paragraph{Classical methods}
The first paper to deal with MVPS is by Hernandez et al. \cite{EstebanVC08}. To avoid having to arbitrate the conflicts between the different normal maps, a 3D mesh is iteratively deformed, starting from the visual hull  
until the images recomputed using the Lambertian model match the original images, while penalizing the discrepancy between the PS normals and those of the 3D mesh. No prior knowledge of camera poses or illumination is required. Under the same assumptions, Park et al. \cite{ParkSMTK13,ParkSMTK17} start from a 3D mesh obtained by SfM (structure-from-motion) and MVS. Simultaneous estimation of reflectance, normals and illumination is achieved by uncalibrated PS, using the normals from the 3D mesh to remove the ambiguity, and estimating the details of the relief through 2D displacement maps. 

MVPS is solved for the first time with a SDF representation of the surface by Logothetis et al. \cite{LogothetisMC19}. Therein, illumination is represented as near point light sources which are assumed calibrated, as well as the camera poses. Thanks to a voxel-based implementation, the surface details are better rendered than with the method of Park et al. \cite{ParkSMTK17}.

Li et al \cite{LiZWSDT20} refine a 3D mesh obtained by propagating the SfM points according to \cite{nehab2005}, and estimate the BRDF using a calibrated setup. The creation of the public dataset ``DiLiGenT-MV'' validates numerically the improved results, in comparison with those of \cite{ParkSMTK17}.
\vspace{-3mm}
\paragraph{Deep learning-based methods}
Kaya et al. \cite{KayaKSFG22} proposed a solution to MVPS based on neural radiance fields (NeRFs)~\cite{nerf}. For each viewpoint, a normal map is obtained using a pre-trained PS network, before a NeRF  is adapted to account for input surface normals from PS in the color function. The recovered geometry yet remains perfectible, according to~\cite{KayaKOFG22}.
Therein, the authors propose learning an SDF function whose zero level set best explains pixel depth and normal maps obtained by a pre-trained MVS~\cite{WangGVSP21} or PS network~\cite{cnnps18}, respectively.
To manage conflicting objectives in the proposed multi-objective optimization and get the best out of MVS and PS predictions, both networks are modified to output uncertainty measures on depth and normal predictions.
The SDF optimization is then carried out while accounting for the inferred uncertainties.

PS-NeRF \cite{YangCCCW22} solves MVPS by jointly estimating the geometry, material and illumination. To this end, the authors propose to regularize the gradient of a UNISURF~\cite{OechsleP021} using the normal maps from PS, while relying on multi-layer perceptrons (MLPs) to explicitly model surface normals, BRDF, illumination, and visibility. These MLPs are optimized based on a shadow-aware differentiable rendering layer. A similar track is followed in~\cite{AsthanaS022}, where NeRFs are combined with a physically-based differentiable renderer.

\begin{figure*}[t]
    \centering
    \includegraphics[width=\textwidth]{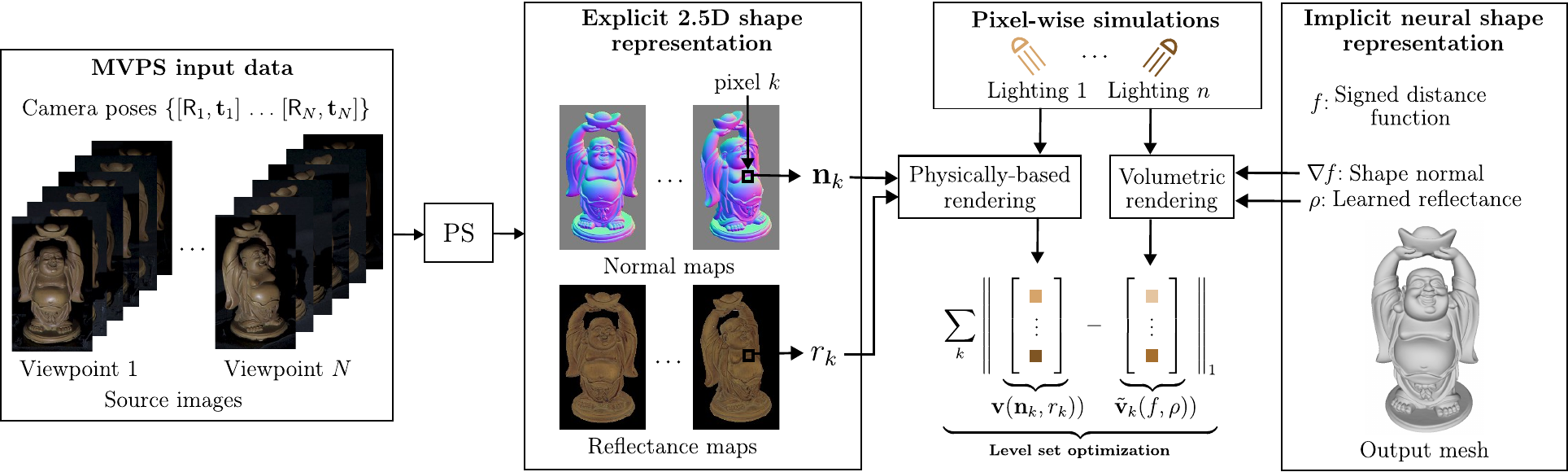}
    \caption{Overview of the proposed MVPS pipeline. The reflectance and normal maps provided for each view by PS are fused, by combining volume rendering with a pixel-wise re-parameterization of the inputs using physically-based rendering.}
    \label{fig:pipeline}
\end{figure*}

Such NeRF-based approaches provide undeniably better 3D reconstructions than classical methods, yet they remain computationally intensive. Recently, Zhao et al.~\cite{mvpsnet} proposed a fast deep learning-based solution to MVPS.
Aggregated shading patterns are matched across viewpoints so that to predict pixel depths and normal maps.

In \cite{KayaKOFG23}, the authors proposed to complement \cite{KayaKOFG22} by adding a NVR loss term in order to benefit from the reliability of NVR in reconstructing objects with diverse material types. However, this results in a multi-objective optimization comprising three loss terms (besides the Eikonal term).
However, similar to \cite{KayaKOFG22},
the uncertainty-based hyper-parameter tuning does not completely eliminate conflicting objectives, which may induce a loss of fine-scale details.
In contrast, we propose a single objective optimization based on an ad hoc re-parametrization which leads to the seamless integration of PS results in standard NVR pipelines. This is detailed in the next paragraph.

%% file: sec/3_method.tex
\section{Proposed approach}
\label{sec:method}

Our aim is to infer a surface whose geometric and photometric properties are consistent with the per-view PS results. To do so, we resort to a volume rendering framework coupled with a re-parameterization of the inputs, as illustrated in Fig.~\ref{fig:pipeline} and 
detailed in the rest of this section.
\subsection{Overview}

\paragraph{Input data} From the $N$ image sets captured under fixed viewpoint and varying illumination, PS provides $N$ reflectance and normal maps, out of which we extract a batch of $m$ posed reflectance and normal values $\{r_k \in \mathbb{R}, \mathbf{n}_k \in \mathbb{S}^2\}_{k=1 \dots m}$. Here, the normal vectors are expressed in world coordinates using the known camera poses. The input reflectance is without loss of generality represented by a scalar (albedo). Let us emphasize that this assumption does not imply that the observed scene must be Lambertian, but rather that we use only the diffuse component of the estimated reflectance. Using other reflectance components (specularity, roughness, etc.), if available, would represent a straightforward extension to more evolved physically-based rendering (PBR) models. Yet, we leave such an extension to perspective for now, since there are few PS methods reliably providing such data. Also, if the PS method provides no reflectance, one can set $r_k \equiv 1$ and use the proposed framework for multi-view normal integration. 

\vspace*{-1em}
\paragraph{Surface parameterization} Our aim is to infer a 3D model of a scene, which consists of both a geometric map $f\!:\mathbb{R}^3 \to \mathbb{R}$ and a photometric one $\rho\!:\mathbb{R}^3 \to \mathbb{R}$. Therein, $f$ associates a 3D point with its signed distance to the surface, which is thus given by the zero level set of $f$: $\mathcal{S}= \{ \mathbf{x} \in \mathbb{R}^3 \,|\, f(\mathbf{x}) = 0\}$. Regarding $\rho$, it encodes the reflectance associated with a 3D point. For input consistency, $\rho$ is considered as a scalar function (albedo), though more advanced PBR models could again be incorporated.

\paragraph{Objective function} Our method builds upon a re-parameterization $\mathbf{v}\!:\mathbb{S}^2 \times \mathbb{R} \to \mathbb{R}^n$ which combines a surface normal $\mathbf{n}_k \in \mathbb{S}^2$ and a reflectance value $r_k \in \mathbb{R}$ into a vector $\mathbf{v}(\mathbf{n}_k,r_k) \in \mathbb{R}^n$ of $n$ radiance values that are simulated by physically-based rendering, using an arbitrary image formation model under varying illumination. Given this re-parameterization, the 3D reconstruction problem amounts to minimizing the difference between a batch of $m$ intensity vectors simulated either from the input data or from volume rendering with the same PBR model, along with a regularization on the SDF: 
\begin{equation}
    \underset{f,\rho}{\min} \sum_{k=1}^m \| \mathbf{v}(\mathbf{n}_k,r_k) - \tilde{\mathbf{v}}_k(f,\rho) \|_1 + \lambda \, \mathcal{L}_\text{reg}(f). 
    \label{eq:photometric_loss}
\end{equation}
Here, $\{(\mathbf{n}_k,r_k)\}_{k=1 \dots m}$ stands for the batch of input reflectance and normal  values, $\mathbf{v}(\mathbf{n}_k,r_k)$ for the $k$-th intensity vector simulated from the input data, $\tilde{\mathbf{v}}_k(f,\rho)$ for the corresponding one simulated by volume rendering, and $\lambda >0$ is a tunable hyper-parameter for balancing the data fidelity with the regularizer $\mathcal{L}_\text{reg}$. The actual optimization can then be carried out seamlessly by resorting to a volume rendering-based 3D reconstruction pipeline such as NeuS~\cite{WangLLTKW21}, given that both $\tilde{\mathbf{v}}_k(f,\rho)$ and $\mathbf{v}(\mathbf{n}_k,r_k)$ correspond to pixel intensities. Let us now detail how we simulate the latter intensities $\mathbf{v}(\mathbf{n}_k,r_k)$ from the input reflectance and normal  data.  

\subsection{Reflectance and normal  re-parameterization}
\label{sec:param}

The input reflectance $\{r_k \in \mathbb{R}\}_k$ and normals $\{\mathbf{n}_k \in \mathbb{S}^2\}_k$ values constitute inhomogeneous quantities: the former are photometric scalars, and the latter geometric vectors lying on the three-dimensional unit sphere. Direct optimization of their consistency with the scene normal $\frac{\nabla f}{\|\nabla f\|}$ and albedo $\rho$ would lead to multiple objectives balanced by hyper-parameters. 

Instead, we propose to jointly re-parameterize the reflectance and normal data into a set of vectors $\{\mathbf{v}(\mathbf{n}_k,r_k) \in \mathbb{R}^n\}_k$ of homogeneous quantities, namely radiance values simulated using a PBR model under varying illumination. In order to enforce the bijectivity of this re-parameterization, we choose as PBR model the linear Lambertian one, under pixel-wise varying illumination represented by $n=3$ arbitrary illumination vectors $\mathbf{l}_{k,1},\mathbf{l}_{k,2},\mathbf{l}_{k,3} \in \mathbb{R}^3$: 
\begin{align}\label{eq:paramA}
\mathbf{v}(\mathbf{n}_k,r_k)  &= r_k [ \mathbf{n}_k^{\top}\mathbf{l}_{k,1}, \mathbf{n}_k^{\top} \mathbf{l}_{k,2}, \mathbf{n}_k^{\top} \mathbf{l}_{k,3}]^{\top}\\
&= r_k \mathsf{L}_k \, \mathbf{n}_{k}, \nonumber
\end{align}
\noindent with $\mathsf{L}_k = \left[\mathbf{l}_{k,1},\mathbf{l}_{k,2},\mathbf{l}_{k,3}\right]^\top$ the arbitrary per-pixel illumination matrix.

For the re-reparameterization to be bijective, the reflectance $r_k$ must be non-null (a basic assumption in photographic 3D vision), and $\mathsf{L}_k$ must be non-singular i.e., the lighting directions must be chosen linearly independent. Then, the original reflectance and normal can be retrieved from the simulated intensities by $r_k = \| \mathsf{L}_k^{-1} \mathbf{v}(\mathbf{n}_k,r_k) \|$ and $\mathbf{n}_k = \frac{\mathsf{L}_k^{-1} \mathbf{v}(\mathbf{n}_k,r_k)}{\| \mathsf{L}_k^{-1} \mathbf{v}(\mathbf{n}_k,r_k) \|}$. Considering $n >3$ illumination vectors and resorting to the pseudo-inverse operator might induce more robustness but at the price of losing bijectivity and thus not entirely relying on the PS inputs. We leave this as a possible future work, which might be particularly interesting when the PS inputs are uncertain, or when considering more evolved PBR models involving additional reflectance clues such as roughness, anisotropy or specularity. 

In practice, the choice of each arbitrary triplet of light directions $\mathbf{l}_{k,1},\mathbf{l}_{k,2},\mathbf{l}_{k,3}$ can be made to minimize the uncertainty on the normal estimate. To this end, the illumination triplet proposed in \cite{Drbohlav05} can be considered. Therein, the authors show that the optimal configuration for three images is vectors that are equally spaced in tilt by $120$ degrees, with a constant slant of $54.74$ degrees (wrt to $\mathbf{n}_k$). 

Let us remark that with the above linear model, it is possible to simulate negative radiance values, when one of the  dot products between the normal and the lighting vectors is negative, which corresponds to self-shadowing. While negative radiance values are obviously non physically plausible, this is not a problem for the proposed re-parameterization, as long as it remains consistent with the NVR strategy, which we are now going to detail. 

\subsection{Volume rendering-based 3D reconstruction}\label{sec:vr}

We now turn our attention to deriving the volume rendering function $\tilde{\mathbf{v}}_k$ arising in Eq.~\eqref{eq:photometric_loss}. The role of this function is to simulate, from the scene geometry $f$ and albedo $\rho$, an intensity vector $\tilde{\mathbf{v}}_k$ which will be compared with the vector $\mathbf{v}_k$ that is simulated from the inputs as described in the previous paragraph.  

Our solution largely takes inspiration from the NeuS method~\cite{WangLLTKW21}, that was initially proposed as a solution to the single-light multi-view 3D surface reconstruction problem. Therein, the rendering function follows a volume rendering scheme which accumulates the colors along the ray corresponding to the $k$-th pixel. Denoting by $\mathbf{o}_k \in \mathbb{R}^3$ the camera center for this observation, and by $\mathbf{d}_{k}$ the corresponding viewing direction, this ray is written $\{\mathbf{x}_{k}(t) = \mathbf{o}_k + t \, \mathbf{d}_{k} \, | \, t \geq 0\}$. By extending the NeuS volume renderer to the multi-illumination scenario, each coefficient $\tilde{v}_{k,l}$ of $\tilde{\mathbf{v}}_k$ is then given,  $\forall l \in \{1,2,3\}$, by:
\begin{equation}\label{eq:render_init}
\tilde{v}_{k,l} = \int^{t_{f}}_{t_{n}}
w(t,f(\mathbf{x}_{k}(t))) \, c_l(\mathbf{x}_{k}(t)) \, \mathrm{d}t,
\end{equation}
where \(t_{n},t_f\) stand for the range bounds over which the colors are accumulated. The weight function $w$ is constructed from the SDF $f$ in order to ensure that it is both occlusion-aware and locally maximal on the zero level set, see~\cite{WangLLTKW21} for details. As for the functions $c_l\!:\mathbb{R}^3  \to \mathbb{R}$, they represent the scene's apparent color. In the original NeuS framework, this color depends not only on the 3D locations, but also on the viewing direction $\mathbf{d}_{k}$, and it is directly optimized along with the SDF $f$. Our case, where the albedo is optimized in lieu of the apparent color, and the illumination varies with the data index $k$ and the illumination index $l$, is however slightly different.  

As a major difference with this prototypical NVR-based 3D reconstruction method, we optimize the SDF $f$ and the surface \textit{albedo} i.e., the scene's intrinsic color $\rho$ rather than its apparent color $c_l$. The dependency upon the viewing direction must thus be removed, in order to ensure consistency with the Lambertian model used for simulating the inputs. More importantly, contrarily to NeuS where the illumination is fixed, each input data $v_{k,l}:= r_k \mathbf{n}_k^\top \mathbf{l}_{k,l}$ is simulated under a different, arbitrary illumination $\mathbf{l}_{k,l}$. For the NVR to produce simulations $\tilde{v}_{k,l}$ matching this input set of intensities, it is necessary to explicitly write the dependency of the apparent color $c_l$ upon the scene's geometry $f$, reflectance $\rho$ and illumination $\mathbf{l}_{k,l}$. Our volume renderer is then still given by Eq.~\eqref{eq:render_init}, but the color of each 3D point must be replaced by: 
\begin{equation}\label{eq:lnorm_final}
    c_{l}(\mathbf{x}_{k}(t)) = \rho(\mathbf{x}_k(t)) \nabla f(\mathbf{x}_{k}(t))^{\top} \mathbf{l}_{k,l},
\end{equation}
where the illumination vectors $\mathbf{l}_{k,l}$ are the same as those in Eq.~\eqref{eq:paramA}. 

Let us remark that the scalar product above corresponds, up to a normalization by $\| \nabla f(\mathbf{x}_{k}(t)) \|$, to the shading. Yet, we do not need to apply this normalization, because the regularization term $\mathcal{L}_\text{reg}(f)$ in~\eqref{eq:photometric_loss} will take care of ensuring the unit length of $\nabla f$. Indeed, as in the original NeuS framework, the SDF is regularized using an eikonal term: 
\begin{equation}\label{eq:eikonal}
\mathcal{L}_{\text{reg}}(f) = \dfrac{\sum_{k=1}^m \int^{t_{f}}_{t_{n}} (\|\nabla f (\mathbf{x}_{k}(t)) \|^2 - 1)^2 \, \mathrm{d}t}{m\left(t_f - t_n\right)}.
\end{equation}
Similarly to the original NeuS, an additional regularization based on object masks can also be utilized for supervision, if such masks are provided.

Plugging~\eqref{eq:lnorm_final} into~\eqref{eq:render_init} yields the definition of our volume renderer accounting for the varying, arbitrary illumination vectors $\mathbf{l}_{k,l}$. Next, plugging ~\eqref{eq:paramA},~\eqref{eq:render_init} and~\eqref{eq:eikonal}
into~\eqref{eq:photometric_loss}, we obtain our objective function, which ensures the consistency between the simulations obtained from the input, and those obtained by volume rendering. It should be emphasized that, besides the eikonal regularization -- which is standard and only serves to ensure the unit-length constraint of the normal, our strategy leads to a single objective optimization formulation for NVR-based 3D surface reconstruction from reflectance and normal  data. 

The discretization of the variational problem~\eqref{eq:photometric_loss} is then achieved exactly as in the original NeuS work~\cite{WangLLTKW21}. It is based on representing $f$ and $\rho$ by MLPs and hierarchically sampling points along the rays. 

\section{Application to MVPS}\label{sec:mvps}

We present a standalone MVPS pipeline that is built on top of the proposed reflectance and normal-based 3D reconstruction method. Our MVPS pipeline includes the following steps:\vspace*{.25em}
\begin{enumerate}
    \item Compute the reflectance and normals maps for each viewpoint through PS;\vspace*{.25em}
    \item Select a batch of the most reliable inputs $\{r_k\}$ and $\{\mathbf{n}_k\}$;\vspace*{.25em}
    \item Scale the reflectance values $\{r_k\}$ across the entire image collection;\vspace*{.25em}
    \item Simulate the radiance values following Eq.~(\ref{eq:paramA}), using a pixel-wise optimal lighting triplet $\mathsf{L}_k$;\vspace*{.25em}
    \item Optimize the loss in Eq. (\ref{eq:photometric_loss}) over the SDF $f$ and albedo~$\rho$;\vspace*{.25em} 
    \item Reconstruct the surface from the SDF. 
\end{enumerate}

\vspace*{-1em}
\paragraph{Step 1: PS-based reflectance and normal  estimation} Any PS method is suitable for obtaining the inputs for each viewpoint. However, not all PS methods actually provide reflectance clues, and not all of them can simultaneously handle non-Lambertian surfaces and unknown, complex illumination. CNN-PS~\cite{cnnps18}, for instance, provides only normals, and for calibrated illumination. For these reasons, we base our MVPS pipeline on the recent transformers-based method SDM-UniPS~\cite{sdmunips}, which exhibits remarkable performance in recovering intricate surface normal maps even when images are captured under unknown, spatially-varying lighting conditions in uncontrolled environments.
As advised by the author of~\cite{sdmunips}, when the number of images is too large for the method to be applied, one can simply take the median of the results over sufficiently many $N_\text{trials}$ random trials, each trial involving the random selection of a few number of images. 

\paragraph{Step 2: Uncertainty evaluation}
To prevent poorly estimated normals from corrupting 3D reconstruction, we discard the less reliable ones. To this end, we use as uncertainty measure the average absolute angular deviation of the normals computed over the $N_\text{trials}$ random trials in Step 1. Pixels associated with an uncertainty measure higher than a threshold ($\tau = 15^\circ$ in our experiments) are excluded from the optimization. Advanced uncertainty metrics, as proposed by Kaya et al.~\cite{KayaKOFG22}, could further refine this process. 

\vspace*{-.5em}

\paragraph{Step 3: Reflectance maps scaling}
The individual reflectance maps computed by PS need to be appropriately scaled. This is because in an uncalibrated setting, the reflectance estimate is relative to both the camera's response, and the incident lighting intensity. Consequently, each reflectance map is estimated only up to a scale factor. To estimate this scale factor, the complete pipeline is first run without using the reflectance maps. This provides pairs of homologous points that are subsequently used to scale the reflectance maps.
Concretely, given a pair of neighboring viewpoints, the ratios of corresponding reflectance values between the two viewpoints are stored, and their median is used to adjust each reflectance map's scale factor. This operation is repeated across the entire viewpoint collection. Note that, if the camera's response and the illumination were known i.e., a calibrated PS method was used in Step~1, then the reflectance would be determined without scale ambiguity and this step could be skipped.

\vspace*{-.5em}

\paragraph{Step 4: Radiance simulation}
To simulate the radiance values, we choose as lighting triplet the one which is optimal, relative to the normal $\mathbf{n}_k$ \cite{Drbohlav05}. The actual formula is provided in the supplementary material.

\vspace*{-.5em}

\paragraph{Step 5: Optimization}
The actual optimization of the loss function is carried out using a straightforward adaptation of the NeuS architecture~\cite{WangLLTKW21}, where viewing direction was removed from the network's input to turn radiance into albedo. 
In all our experiments, we let the optimization run for a total of 300k iterations, with a batch size of 512 pixels. To ensure that the networks have a better understanding of our MVPS data, we decided to train each iteration not only on a random view, but also on all rendered images of this view under varying illumination. The backward operation is then applied only after the loss is computed on all pixels for all the illumination conditions. In terms of computation time, our approach is comparable with the original NeuS framework, requiring in our tests from $8$ to $16$ hours on a standard GPU for the 3D reconstruction of each dataset from DiLiGenT-MV~\cite{LiZWSDT20}.

\vspace*{-.5em}

\paragraph{Step 6: Surface reconstruction} Once the SDF is estimated, we extract its zero level set using the marching cube algorithm~\cite{LorensenC87}.

%% file: sec/4_experiments.tex
\section{Experimental results}
\label{sec:xp}

\subsection{Experimental setup}

\paragraph{Evaluation datasets}
We used the DiLiGenT-MV benchmark dataset \cite{LiZWSDT20} to perform all our experiments, statistical evaluations, and ablations. It includes five real-world objects with complex reflectance properties and surface profiles, making it an ideal choice for the proposed method evaluation. Each object is imaged from 20 calibrated viewpoints using the classical turntable MVPS acquisition setup~\cite{EstebanVC08}. For each view, 96 images are acquired under different illuminations. Given the large volume of images, which is impractical for transformers-based methods, our implementation of Step 1 (PS) employs SDM-UniPS~\cite{sdmunips} with only $10$ input images. To this end, we computed each $r_k$ and $\mathbf{n}_k$ as the medians of the computed reflectances and normals over $N_\text{trials} = 100$ random trials, each trial involving the random selection of 10 images from the 96 available in the DiLiGenT-MV dataset.

\paragraph{Evaluation scores}
We performed our quantitative evaluations using F-score and Chamfer distance (CD), to measure the accuracy of the reconstructed vertices. We also measured the mean angular error (MAE) of the imaged meshes, to evaluate the accuracy of the reconstructed normals wrt the ground truth normals provided in DiLiGenT-MV.
We report both the results averaged over all mesh vertices, and those on vertices clustered in two particularly interesting classes, namely high curvature and low visibility areas, as illustrated in Fig.~\ref{fig:visibility_seg}.

\input{figures/visibility_binaire} 

To identify the high curvature areas, we used the library VCGLib \cite{vcglib} and the 3D mesh processing software system Meshlab \cite{meshlab}, taking the absolute value of the curvature to merge the convex and concave zones and retaining the vertices whose curvature is higher than $1.6$. To segment the low visibility areas, we summed the boolean visibility of each vertex in each view. Low visibility then corresponds to vertices visible in less than 5 viewpoints, among the 20 ones of DiLiGenT-MV.  

\input{tables/table_mesh_mae}

\subsection{Baseline comparisons}

We first provide in Fig.~\ref{fig:mesh_mae} a qualitative comparison of our results on four objects, and compare them with the three most recent methods from the literature, namely PS-NERF~\cite{YangCCCW22}, Kaya23~\cite{KayaKOFG23} and MVPSNet~\cite{mvpsnet}. In comparison with these state-of-the-art deep learning-based methods, the recovered geometry is overall more satisfactory. 

This is confirmed quantitatively when evaluating Chamfer distances and MAE, provided in Tables~\ref{table:chamfer_global} and~\ref{table:mae_global}. Therein, beside the aforementioned methods we also report the results from the Kaya22 method~\cite{KayaKOFG22} and those from the non deep learning-based ones Park16~\cite{ParkSMTK17} and Li19~\cite{LiZWSDT20} (which is not fully automatic). From the tables, it can be seen that our method outperforms other fully automated standalone ones, and is competitive with the semi-automated one. On average, our method reports a Chamfer distance which is $17.4\%$ better than the second best score, obtained by MVPSNet~\cite{mvpsnet}. Regarding MAE, our  score is similar to Kaya23~\cite{KayaKOFG23} with a small average difference of 0.2 degree. The superiority of our approach can also be observed by considering the F-scores, which are reported in Fig.~\ref{fig:fscore}. 

\input{tables/table_CD_global} 
\input{tables/table_MAE_global} 
\input{figures/figure_fscore_standalone_ablation} 

\subsection{High curvature and low visibility areas}

To highlight the level of details in the 3D reconstructions, Figs.~\ref{fig:example} and~\ref{fig:cmp_zoom} provide other qualitative comparisons focusing on one small part of each object. Ours is the only method achieving a high fidelity reconstruction on the ear, the knot and the navel of Buddha, and on the spout of Pot2. To quantify this gain, we also report in Table~\ref{table:table_CCHV} the average CD and MAE over all datasets, yet taking into account only the high curvature and low visibility areas. It is worth noticing that the CD error of PS-NeRF and MVPSNet on high curvature areas increases by $36\%$ and $96\%$, respectively, in comparison with that averaged over the entire set of vertices. Ours, on the contrary, increases by $4\%$ only. Similarly, on low visibility areas their error increases by $78\%$ and $81\%$, and Kaya23 by $46\%$, while ours increases only by $13\%$. 

\input{tables/table_curv_vis}
\input{figures/cmp_methods_zoom} 

\subsection{Ablation study}

Lastly, we conducted an ablation study, to quantify the impact of some parts of our pipeline. More precisely, we quantify in Fig.~\ref{fig:fscore}\textcolor{red}{b} and Table~\ref{table:chamfer_ablation} the impact of providing PS-estimated reflectance maps, in comparison with providing only normals (``W/o reflectance''). We also evaluate that of the pixel-wise optimal lighting triplet, in comparison with using the same arbitrary one for all pixels in one view (``W/o optimal lighting''). Lastly, we evaluate the impact of discarding the less reliable inputs, in comparison with using all of them (``W/o uncertainty''). The feature that influences most the accuracy of the 3D reconstruction is the use of reflectance.
The other two features also positively impact the reconstruction, but to a lesser extent. 

\input{tables/table_CD_ablation}

\subsection{Limitations}

Our approach heavily relies on the quality of the PS normal maps. In our experiments, we used  SDM-UniPS~\cite{sdmunips}, which generally yields high quality results. Yet, it occasionally yields corrupted normals, leading to inconsistencies across viewpoints that may result in errors in the reconstruction (cf. supplementary material). This could be handled in the future by replacing the PS method by a more robust one. A second limitation, similar to PS-NeRF, is the computation time, which falls within the range of 8 to 16 hours for one object in DiLiGenT-MV. Fortunately, NeuS2~\cite{neus2}, a significantly faster version of NeuS, will allow us to reduce the computation time to around ten minutes. 

%% file: figures/visibility_binaire.tex
\begin{figure}[!ht]
\centering
\begin{tabular}{cc}
\hspace{-3mm}
\includegraphics[height = 0.5\linewidth]{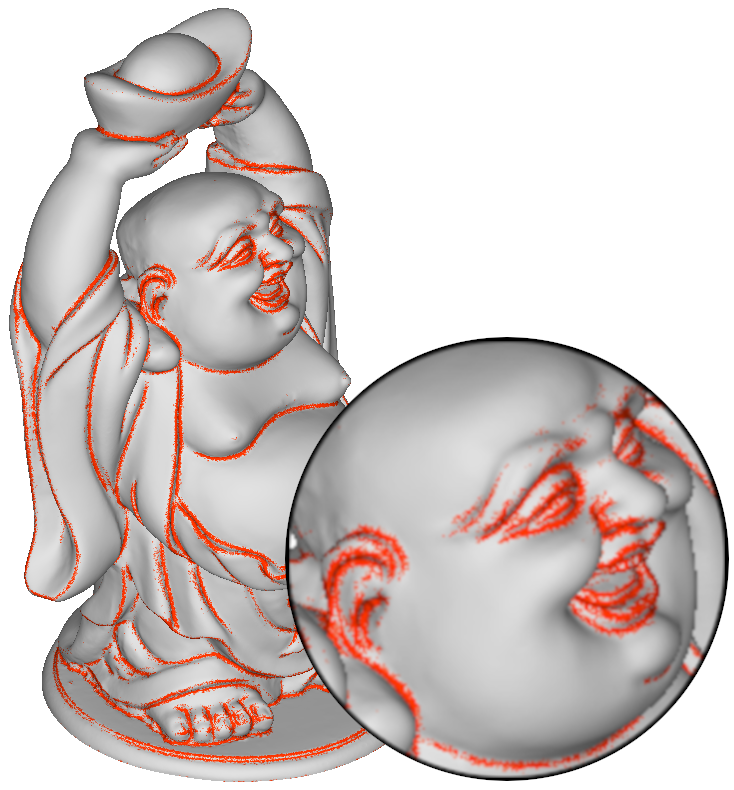} &
\includegraphics[height = 0.42\linewidth]{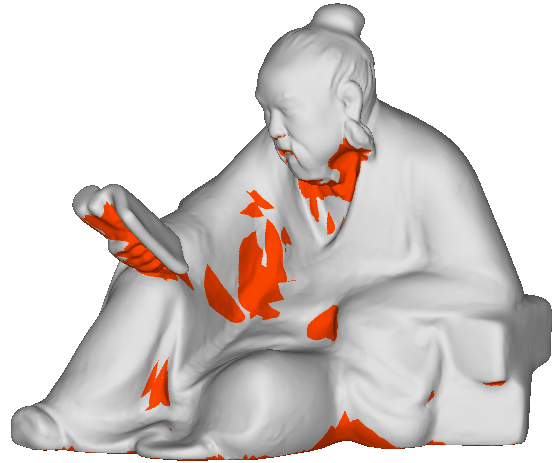} 
\end{tabular}
\caption{High curvature (left) and low visibility (right) areas, on the Buddha and Reading datasets.}
\label{fig:visibility_seg}
\end{figure}

%% file: tables/table_mesh_mae.tex
\begin{figure*}[h!]
\hspace{-3mm}
\begin{tabular}{ccc}
\includegraphics[width = 0.48\linewidth]{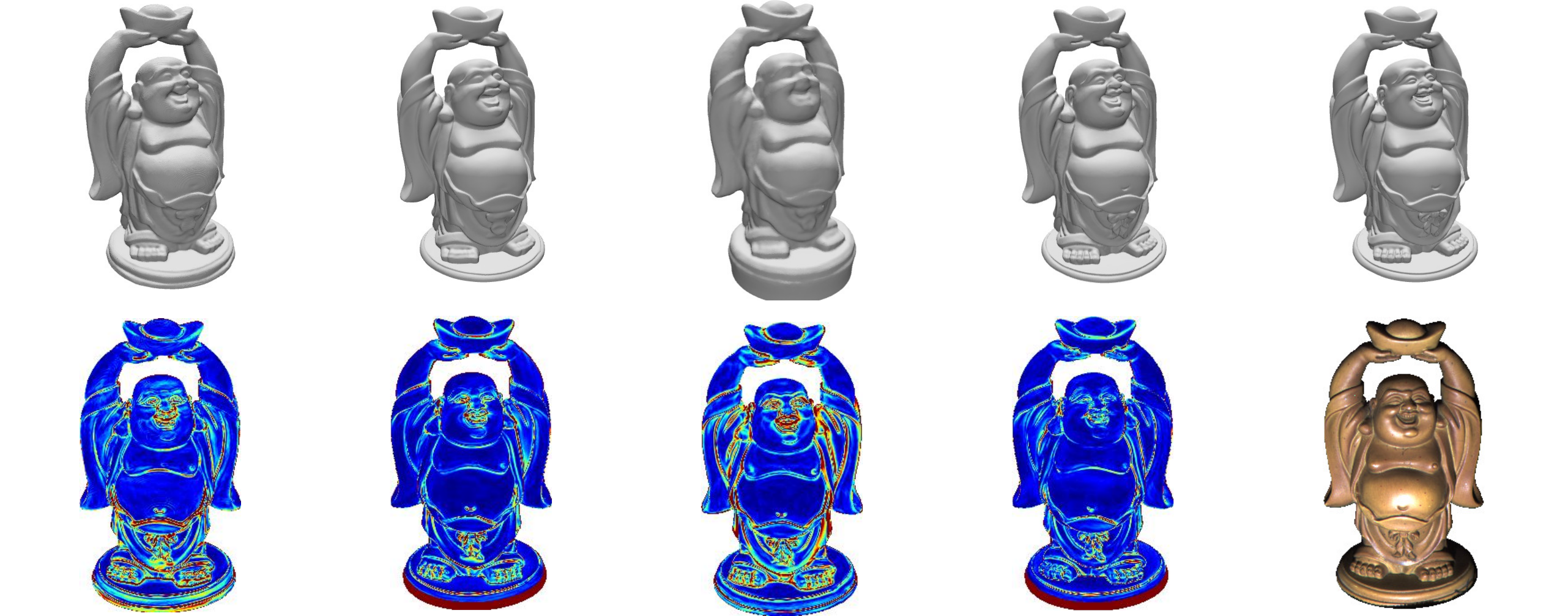} & &
\includegraphics[width = 0.48\linewidth]{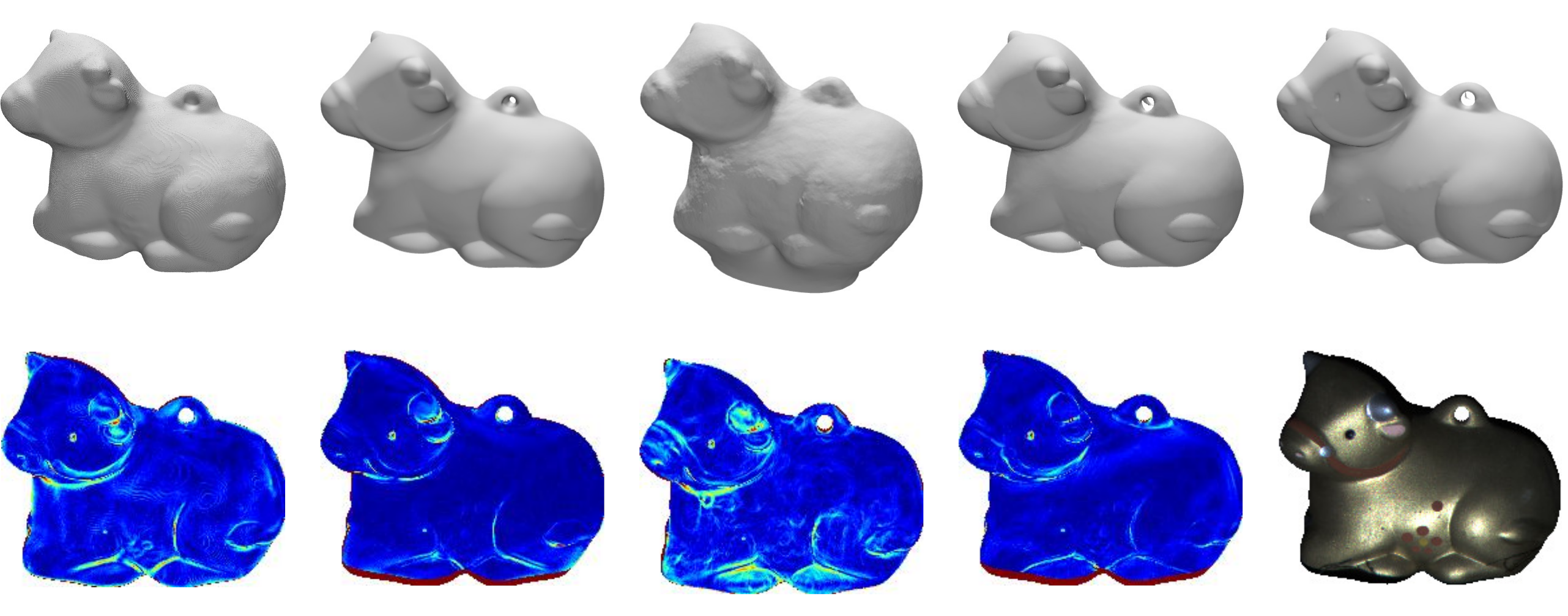} \\
~~{\small PS-NeRF} ~~~~~ {\small Kaya23} ~~~~ {\small MVPSNet} ~~~~~~ {\small Ours} ~~~~~~~~~~~ {\small GT}~~~~~~~  &  &  
~~~~{\small PS-NeRF} ~~~~~ {\small Kaya23} ~~~~~ {\small MVPSNet} ~~~~~~ {\small Ours} ~~~~~~~~~~~ {\small GT}~~~~~~~ \vspace{-1.7mm} \\
\vspace{-1.2mm}
---------------------------------------------------------------- & & ----------------------------------------------------------------\\
{\small Buddha} & & {\small Cow} \\
\includegraphics[width = 0.48\linewidth]{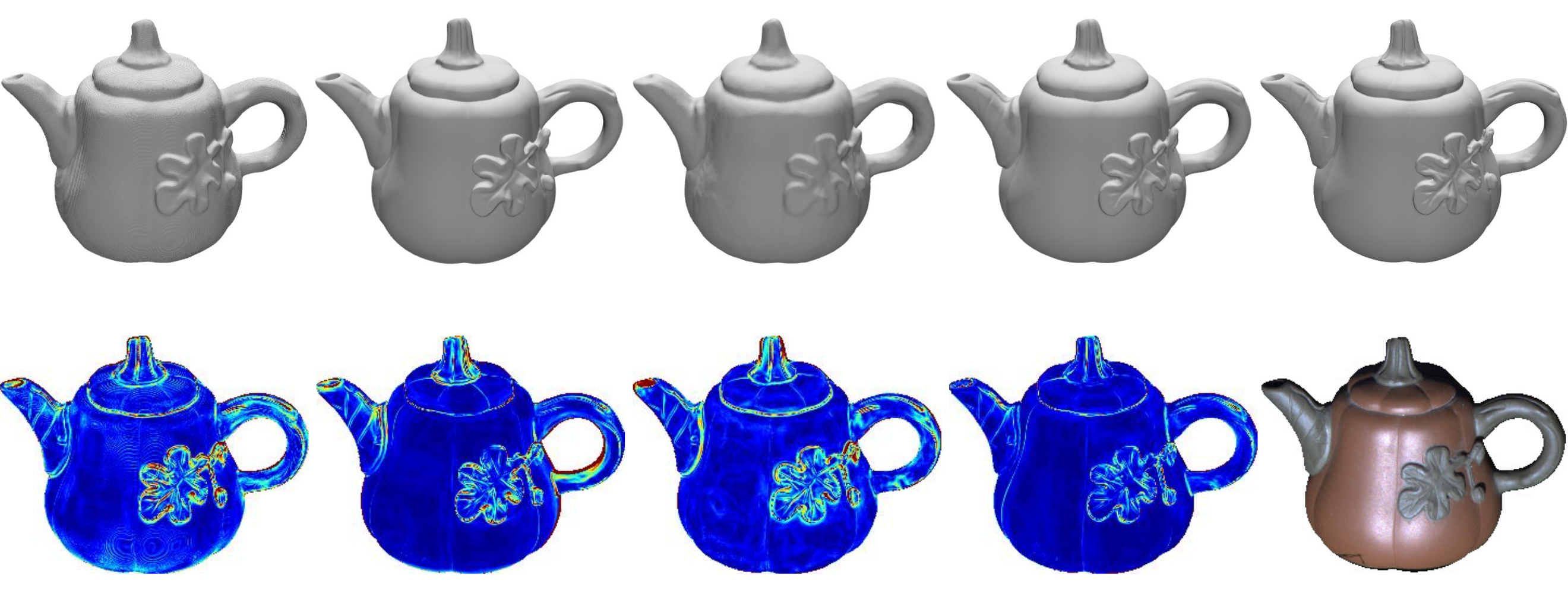} & &
\includegraphics[width = 0.48\linewidth]{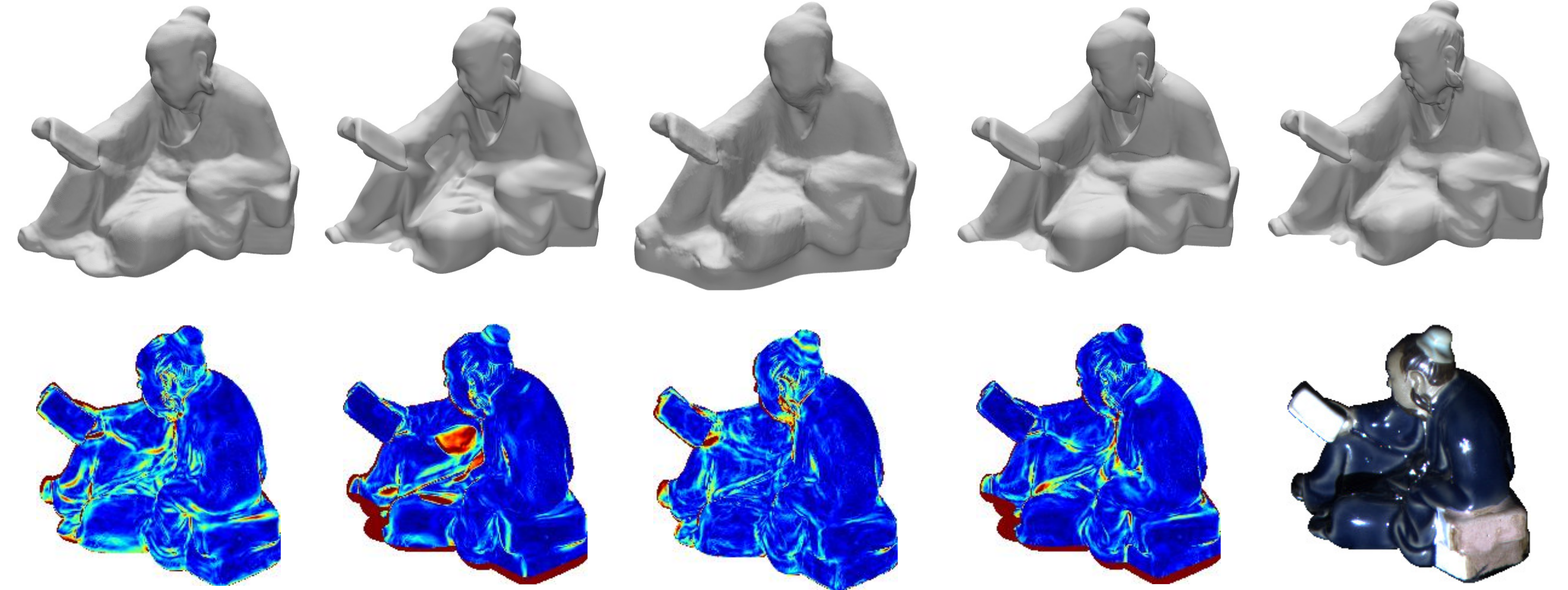} \\
~~{\small PS-NeRF} ~~~~~ {\small Kaya23} ~~~~ {\small MVPSNet} ~~~~~~ {\small Ours} ~~~~~~~~~~~ {\small GT}~~~~~~~  &  &  
~~~~{\small PS-NeRF} ~~~~~ {\small Kaya23} ~~~~~ {\small MVPSNet} ~~~~~~ {\small Ours} ~~~~~~~~~~~ {\small GT}~~~~~~~ \vspace{-1.7mm} \\
\vspace{-1.2mm}
---------------------------------------------------------------- & & ----------------------------------------------------------------\\
{\small Pot2} & & {\small Reading} \vspace*{-.5em}
\end{tabular}
\caption{Reconstructed 3D mesh and corresponding angular error of four objects from the DiLiGenT-MV benchmark.\vspace*{-1em} }
\label{fig:mesh_mae}
\end{figure*}

%% file: tables/table_CD_global.tex
\begin{table}[h!]
\centering
\small
\begin{tabular}{l|ccccc|c}
& \multicolumn{6}{c}{Chamfer distance $\downarrow$}\\
Methods & Bear & Budd. & Cow & Pot2 & Read. & Aver. \\
\hline
Park16 &  0.92 & 0.39 & 0.34 & 0.94 & 0.53 & 0.62\\
Li19 $\dagger$ &  0.22 & 0.28 & 0.11 & 0.23 & 0.27 & 0.22\\
Kaya22 &  0.39 & 0.4 & 0.3 & 0.4 & 0.35 & 0.37\\
PS-NeRF &  0.32 & 0.28 & \textcolor{bestblue}{\textbf{0.24}} & \textcolor{bestblue}{\textbf{0.24}} & 0.33 & 0.28\\
Kaya23 &  0.33 & \textcolor{bestgreen}{\textbf{0.21}} & \textcolor{bestgreen}{\textbf{0.22}} & 0.37 & 0.28 & 0.28\\
MVPSNet &  \textcolor{bestblue}{\textbf{0.28}} & 0.3 & 0.25 & 0.27 & \textcolor{bestgreen}{\textbf{0.25}} & \textcolor{bestblue}{\textbf{0.27}}\\
Ours &  \textcolor{bestgreen}{\textbf{0.22}} & \textcolor{bestblue}{\textbf{0.22}} & 0.25 & \textcolor{bestgreen}{\textbf{0.16}} & \textcolor{bestblue}{\textbf{0.27}} & \textcolor{bestgreen}{\textbf{0.23}}\\
\end{tabular}
\caption{Chamfer distance (lower is better) averaged overall all vertices. \textcolor{bestgreen}{\textbf{Best results}}. \textcolor{bestblue}{\textbf{Second best}}. Since $\dagger$ requires manual efforts, it is not ranked.}
\label{table:chamfer_global}
\end{table}

%% file: tables/table_MAE_global.tex
\begin{table}[h!]
\centering
\small
\begin{tabular}{l|ccccc|c}
& \multicolumn{6}{c}{Normal MAE $\downarrow$}\\
Methods & Bear & Budd. & Cow & Pot2 & Read. & Aver. \\
\hline
Park16 &  9.64 & 12.6 & 8.23 & 11.1 & 9.01 & 10.1\\
Li19 $\dagger$ &  3.85 & 11.0 & 2.82 & 5.88 & 6.30 & 5.97\\
Kaya22 &  4.89 & 12.5 & 4.44 & 8.68 & 6.52 & 7.41\\
PS-NeRF &  5.48 & 11.7 & 5.46 & 7.65 & 9.13 & 7.88\\
Kaya23 &  \textcolor{bestblue}{\textbf{3.24}} & \textcolor{bestgreen}{\textbf{8.12}} & \textcolor{bestgreen}{\textbf{3.04}} & 5.63 & \textcolor{bestgreen}{\textbf{5.66}} & \textcolor{bestblue}{\textbf{5.14}}\\
MVPSNet &  5.26 & 14.1 & 6.28 & 6.69 & 8.58 & 8.18\\
SDM-UniPS* &  4.79 & 9.60 & 5.46 & \textcolor{bestblue}{\textbf{5.56}} & 10.1 & 7.12\\
Ours &  \textcolor{bestgreen}{\textbf{2.70}} & \textcolor{bestblue}{\textbf{8.17}} & \textcolor{bestblue}{\textbf{3.61}} & \textcolor{bestgreen}{\textbf{4.11}} & \textcolor{bestblue}{\textbf{6.18}} & \textcolor{bestgreen}{\textbf{4.95}}\\
\end{tabular}
\caption{Normal MAE (lower is better) averaged over all views. For reference, the mono-view PS results from SDM-UniPS~\cite{sdmunips} (*) are also provided, although it does not provide a full 3D reconstruction and thus its Chamfer distance cannot be evaluated.}
\label{table:mae_global}
\end{table}

%% file: figures/figure_fscore_standalone_ablation.tex
\begin{figure}[h]
\centering
\begin{tabular}{cc}
\hspace{-3mm}\includegraphics[width = 0.495\linewidth]{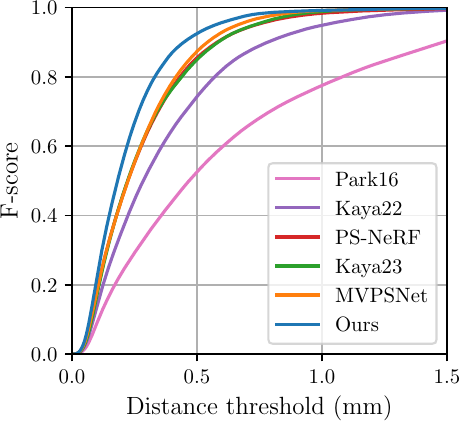} &
\hspace{-2.5mm}\includegraphics[width = 0.505\linewidth]{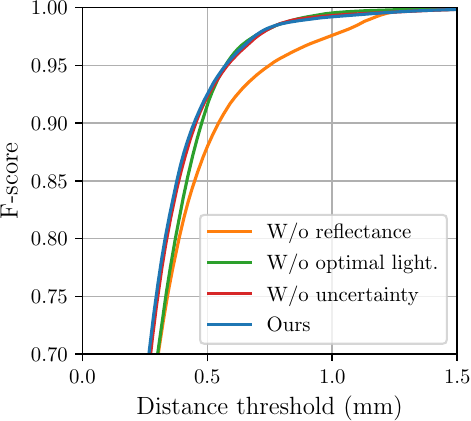} \\
{\small (a)} & {\small (b)} \vspace*{-.5em}
\end{tabular}
\caption{F-score (higher is better) as a function
of the distance error threshold, in comparison with other state-of-the-art methods (a), and disabling individual components of our method (b). 
}
\label{fig:fscore}
\end{figure}

%% file: tables/table_curv_vis.tex
\begin{table}[h!]
\centering
\small
\begin{tabular}{l|c|c|c|c|c|c}
& \multicolumn{2}{c|}{\textbf{All}} &  \multicolumn{2}{c|}{\textbf{High curv.}} & \multicolumn{2}{c}{\textbf{Low vis.}}\\
\hline
\% Vertices & \multicolumn{2}{c|}{100\%} & \multicolumn{2}{c|}{8.27\%} & \multicolumn{2}{c}{8.70\%}\\
\hline
Scores & CD & MAE  & CD & MAE & CD & MAE\\
\hline
Park16 &  0.62 & 10.1 & 0.88 & 29.0 & 0.68 & 29.6 \\
Li19 $\dagger$ &  0.22 & 5.97 & 0.51 & 26.2 & 0.67 & 33.3 \\
Kaya22 &  0.37 & 7.41 & 0.45 & 28.0 & 0.54 & 31.7 \\
PS-NeRF &  0.28 & 7.88 & 0.38 & 25.8 & 0.5 & 24.0 \\
Kaya23 &  0.28 & \textcolor{bestblue}{\textbf{5.14}} & \textcolor{bestblue}{\textbf{0.29}} & \textcolor{bestblue}{\textbf{23.6}} & \textcolor{bestblue}{\textbf{0.41}} & \textcolor{bestblue}{\textbf{20.7}} \\
MVPSNet &  \textcolor{bestblue}{\textbf{0.27}} & 8.18 & 0.53 & 23.9 & 0.49 & 28.9 \\
Ours &  \textcolor{bestgreen}{\textbf{0.23}} & \textcolor{bestgreen}{\textbf{4.95}} & \textcolor{bestgreen}{\textbf{0.24}} & \textcolor{bestgreen}{\textbf{23.1}} & \textcolor{bestgreen}{\textbf{0.26}} & \textcolor{bestgreen}{\textbf{17.8}} 
\end{tabular}
\caption{Chamfer distance and normal MAE (lower is better) on high curvature and low visibility areas.}
\label{table:table_CCHV}
\end{table}

%% file: figures/cmp_methods_zoom.tex
\begin{figure*}[!ht]
\centering
\def\svgwidth{\linewidth}
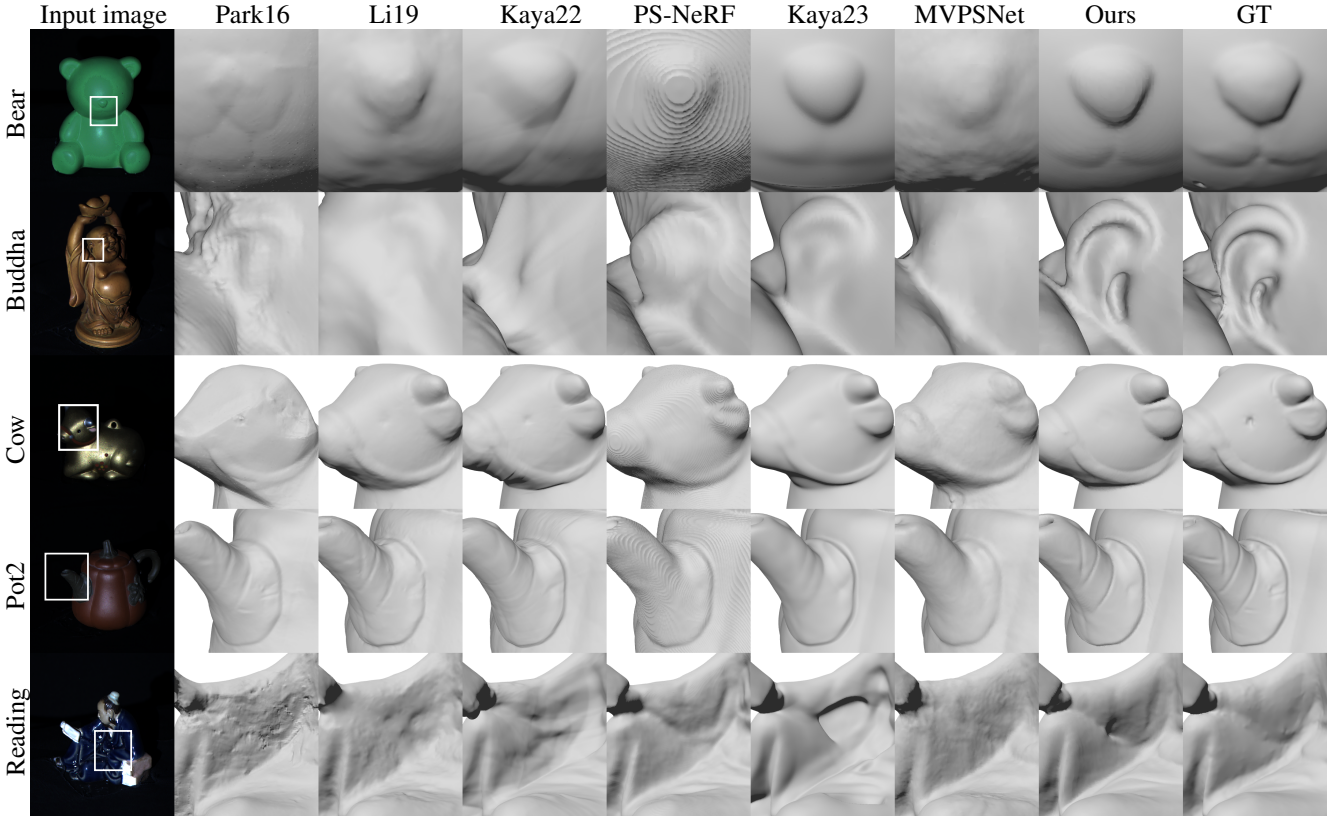

\caption{Qualitative comparison between our results and state-of-the-art ones, on parts of the meshes representing fine details. 
}
\label{fig:cmp_zoom}
\end{figure*}

%% file: figures/image_all_methods_V2.pdf_tex
\begingroup%
  \makeatletter%
  \providecommand\color[2][]{%
    \errmessage{(Inkscape) Color is used for the text in Inkscape, but the package 'color.sty' is not loaded}%
    \renewcommand\color[2][]{}%
  }%
  \providecommand\transparent[1]{%
    \errmessage{(Inkscape) Transparency is used (non-zero) for the text in Inkscape, but the package 'transparent.sty' is not loaded}%
    \renewcommand\transparent[1]{}%
  }%
  \providecommand\rotatebox[2]{#2}%
  \newcommand*\fsize{\dimexpr\f@size pt\relax}%
  \newcommand*\lineheight[1]{\fontsize{\fsize}{#1\fsize}\selectfont}%
  \ifx\svgwidth\undefined%
    \setlength{\unitlength}{3093.55428343bp}%
    \ifx\svgscale\undefined%
      \relax%
    \else%
      \setlength{\unitlength}{\unitlength * \real{\svgscale}}%
    \fi%
  \else%
    \setlength{\unitlength}{\svgwidth}%
  \fi%
  \global\let\svgwidth\undefined%
  \global\let\svgscale\undefined%
  \makeatother%
  \begin{picture}(1,0.61177569)%
    \lineheight{1}%
    \setlength\tabcolsep{0pt}%
    \put(0,0){\includegraphics[width=\unitlength,page=1]{image_all_methods_V2.pdf}}%
    \put(0.00868571,0.51572696){\rotatebox{90.905194}{\makebox(0,0)[lt]{\lineheight{1.25}\smash{\begin{tabular}[t]{l}$\text{Bear}$\end{tabular}}}}}%
    \put(0.00843636,0.15670416){\rotatebox{90.905194}{\makebox(0,0)[lt]{\lineheight{1.25}\smash{\begin{tabular}[t]{l}$\text{Pot2}$\end{tabular}}}}}%
    \put(0.02077216,0.6044078){\makebox(0,0)[lt]{\lineheight{1.25}\smash{\begin{tabular}[t]{l}$\text{Input image}$\end{tabular}}}}%
    \put(0.15429845,0.6044078){\makebox(0,0)[lt]{\lineheight{1.25}\smash{\begin{tabular}[t]{l}$\text{Park16}$\end{tabular}}}}%
    \put(0.27108227,0.6044078){\makebox(0,0)[lt]{\lineheight{1.25}\smash{\begin{tabular}[t]{l}$\text{Li19}$\end{tabular}}}}%
    \put(0.47258047,0.6044078){\makebox(0,0)[lt]{\lineheight{1.25}\smash{\begin{tabular}[t]{l}$\text{PS-NeRF}$\end{tabular}}}}%
    \put(0.37057384,0.6044078){\makebox(0,0)[lt]{\lineheight{1.25}\smash{\begin{tabular}[t]{l}$\text{Kaya22}$\end{tabular}}}}%
    \put(0.58937399,0.6044078){\makebox(0,0)[lt]{\lineheight{1.25}\smash{\begin{tabular}[t]{l}$\text{Kaya23}$\end{tabular}}}}%
    \put(0.68638377,0.6044078){\makebox(0,0)[lt]{\lineheight{1.25}\smash{\begin{tabular}[t]{l}$\text{MVPSNet}$\end{tabular}}}}%
    \put(0.81625698,0.6044078){\makebox(0,0)[lt]{\lineheight{1.25}\smash{\begin{tabular}[t]{l}$\text{Ours}$\end{tabular}}}}%
    \put(0.93170788,0.6044078){\makebox(0,0)[lt]{\lineheight{1.25}\smash{\begin{tabular}[t]{l}$\text{GT}$\end{tabular}}}}%
    \put(0.00868579,0.38397015){\rotatebox{90.905194}{\makebox(0,0)[lt]{\lineheight{1.25}\smash{\begin{tabular}[t]{l}$\text{Buddha}$\end{tabular}}}}}%
    \put(0.00842413,0.26807428){\rotatebox{90.905194}{\makebox(0,0)[lt]{\lineheight{1.25}\smash{\begin{tabular}[t]{l}$\text{Cow}$\end{tabular}}}}}%
    \put(0.00872005,0.03034817){\rotatebox{90.905194}{\makebox(0,0)[lt]{\lineheight{1.25}\smash{\begin{tabular}[t]{l}$\text{Reading}$\end{tabular}}}}}%
  \end{picture}%
\endgroup%

%% file: tables/table_CD_ablation.tex
\begin{table}[h!]
\centering
\small
\begin{tabular}{l|ccccc|c}
& \multicolumn{6}{c}{Chamfer distance $\downarrow$}\\
Methods & Bear & Budd. & Cow & Pot2 & Read. & Aver. \\
\hline
W/o reflect. &  0.23 & 0.22 & 0.39 & 0.16 & 0.31 & 0.26 \\
W/o opt. l. &  0.32 & \textcolor{bestgreen}{\textbf{0.22}} & \textcolor{bestgreen}{\textbf{0.20}} & 0.19 & \textcolor{bestgreen}{\textbf{0.27}} & 0.24 \\
W/o uncert. &  0.22 & 0.22 & 0.27 & 0.16 & 0.27 & 0.23 \\
Ours &  \textcolor{bestgreen}{\textbf{0.22}} & 0.22 & 0.25 & \textcolor{bestgreen}{\textbf{0.16}} & 0.27 & \textcolor{bestgreen}{\textbf{0.23}} \\
\end{tabular}
\caption{Chamfer distance (lower is better) averaged overall all vertices, while disabling individual features of the pipeline (reflectance estimation, optimal lighting, and uncertainty evaluation).}
\label{table:chamfer_ablation}
\end{table}

%% file: sec/5_conclusion.tex
\section{Conclusion}
\label{sec:concl}

We have introduced a neural volumetric rendering method for 3D surface reconstruction based on reflectance and normal maps, and applied it to multi-view photometric stereo. The proposed method relies on a joint re-parameterization of reflectance and normal as a vector of radiances rendered under simulated, varying illumination. It involves a single objective optimization, and it is highly flexible since any existing or future PS method can be used for constructing the input reflectance and normal maps.
Coupled with a state-of-the-art uncalibrated PS method, our method reaches unprecedented results on the public dataset DiLiGenT-MV in terms of F-score, Chamfer distance and mean angular error metrics. Notably, it provides exceptionally high quality results in areas with high curvature or low visibility. 
Its main limitation for now is its computational cost, which we plan to reduce by adapting recent developments within the NeuS2 framework~\cite{neus2}.
Using reflectance uncertainty in addition to that of normal maps offers room for improvement.
\vspace{-8mm}
\paragraph{Acknowledgements.}
This work was supported by the Danish project PHYLORAMA, the ALICIA-Vision project, the IMG project (ANR-20-CE38-0007), the OR-X and associated funding by the University of Zurich and University Hospital Balgrist. 


%% file: sec/X_suppl.tex

\newpage

{\Large
\noindent \textbf{Appendix}}\\
\vspace{1.0em}
\appendix

This supplementary material provides technicalities and detailed analysis of the experiments. We provide the reader with explicit formulations of the evaluation metrics in Section \ref{sec:metrics}. We then share additional implementation details in Section \ref{sec:impl}. In Section \ref{sec:results}, we present additional quantitative and qualitative results. In Section \ref{sec:limits}, we illustrate some limitations of our method.

\section{Evaluation}
\label{sec:metrics}

\paragraph{Metrics.}
All quantitative evaluations were carried out using Chamfer distance, F-score and mean angular error (MAE) between the reconstructed mesh $\mathcal{P}$ and the ground truth one $\mathcal{G}$.
For a reconstructed point $\hat{\mathbf{x}} \in \mathcal{P}$, its distance to the ground truth is defined
as follows:
\begin{equation}
    d_{\hat{\mathbf{x}}\rightarrow \mathcal{G}} = \underset{\mathbf{x}\in \mathcal{G}}{\operatorname{min}}
\| \hat{\mathbf{x}} - \mathbf{x} \|,
\end{equation}
and vice versa for a ground truth point $\mathbf{x} \in \mathcal{G}$ and its distance to the reconstructed mesh.

The distance measures are accumulated over the entire meshes to define the Chamfer distance 
\begin{equation}
    CD =
    \frac{1}{2}\left ( \frac{1}{|\mathcal{P}|} \sum_{\hat{\mathbf{x}} \in \mathcal{P}} d_{\hat{\mathbf{x}}\rightarrow \mathcal{G}}
    +
    \frac{1}{|\mathcal{G}|} \sum_{\mathbf{x} \in \mathcal{G}} d_{\mathbf{x}\rightarrow \mathcal{P}} \right )
\end{equation}
and the F-score
\begin{equation}
    F(\epsilon) = \frac{2 P(\epsilon) R(\epsilon)}{P(\epsilon) + R(\epsilon)},
\end{equation}
where 
\begin{equation}
    P(\epsilon) = \frac{1}{|\mathcal{P}|} \sum_{\hat{\mathbf{x}} \in \mathcal{P}}
    [ d_{\hat{\mathbf{x}}\rightarrow \mathcal{G}} < \epsilon ]
\end{equation}
and
\begin{equation}
    R(\epsilon) = \frac{1}{|\mathcal{G}|} \sum_{\mathbf{x} \in \mathcal{G}}
    [ d_{\mathbf{x}\rightarrow \mathcal{P}} < \epsilon ]
\end{equation}
are precision and recall measures, respectively, $[.]$ is the Iverson bracket and $\epsilon$ is the distance threshold.

The mesh segmentations into low visibility and high curvature areas are performed on the ground truth meshes. Because the geometry of the reconstruction differs from that of the ground truth, the segmentation procedure yields different areas when applied to the reconstruction. For this reason, the reported results for low visibility and high curvature areas only consider the Chamfer distance term indicating the average distances between the ground truth vertices and their nearest neighbors in the reconstructed mesh.

For the MAE computation, the reconstructed and ground truth meshes are projected onto image planes and the normals are computed at each pixel.
The MAE over all the pixels $M$ is written as
\begin{equation}
    MAE = \frac{1}{|M|} \sum_{k\in M} \cos^{-1}({\hat{\mathbf{n}}_k}^\top\mathbf{n}_k).
\end{equation}

\paragraph{DiLiGenT-MV dataset.}
All the state-of-the-art methods were evaluated from the meshes that were kindly provided by their authors.
For all evaluated meshes, we eliminated all internal vertices.
Then, a mesh upsampling for both estimated and ground truth meshes was then performed in order to achieve a point density of $0.1$~mm.
The computations of Chamfer distance and F-score were specifically conducted for distances under $5$~mm in order to mitigate the impact of outliers (inspired by the DTU evaluation~\cite{jensen2014large}).
We observed a few defects in the ground truth meshes from the DiLiGenT-MV dataset in concave areas.
Notably, such imperfections are well visible at the back of Bear's head (Fig.~\ref{fig:bearheat})
and the spout's inner area of Pot2 (Fig.~\ref{fig:pot2section}). 
Although these areas represent a small amount of vertices, they were discarded in all evaluations so as to avoid penalizing methods which faithfully reconstruct them.

\input{figures/bear_eval_masking}
\input{figures/pot2_section}

\paragraph{Manual efforts in~\cite{LiZWSDT20}.} 
Li19 \cite{LiZWSDT20} is mentioned as 
requiring manual efforts. Indeed, the authors manually establish point correspondences in textureless areas. See \cite{LiZWSDT20} for details.

\section{Implementation details}
\label{sec:impl}

We recall that to simulate the radiance values in Step 4 described in Section \textcolor{red}{4} of the main paper, we choose as lighting triplet the one which is optimal, relatively to the normal $\mathbf{n}_k$.
Following \cite{Drbohlav05}, this optimal triplet is equally spaced in tilt 120 degrees apart with a slant angle of 54.74 degrees. 
Concretely, the expression of $\mathsf{L}_k$ as a function of $\mathbf{n}_k$ is written:
\begin{equation}
    \mathsf{L}_k = \mathsf{R}_k \mathsf{L}_{\text{canonic}}
\end{equation}
where $\mathsf{R}_k = \mathsf{U}$ with $[\mathsf{U},\Sigma,\mathsf{U}]=\operatorname{SVD}(\mathbf{n}_k \mathbf{n}_k^{\top})$ and 
\begin{equation}
    \mathsf{L}_{\text{canonic}} = \begin{bmatrix}
\sin(\phi) & \sin(\phi)\cos(\theta) & \sin(\phi)\cos(2\theta) \\
0 & \sin(\phi)\sin(\theta) & \sin(\phi)\sin(2\theta) \\
\cos(\phi) & \cos(\phi) & \cos(\phi) \\
\end{bmatrix}
\end{equation}
with $\theta=\frac{120\pi}{180}$ and $\phi=\frac{54.74\pi}{180}$.

\section{Additional Results}
\label{sec:results}

In this section, we extend the experiments of
the main paper by providing further statistical analysis and qualitative comparisons.

\paragraph{Comparison with mono-illumination NeuS. }
We propose an additional comparison of our method against the multi-view mono-illumination 3D reconstruction method NeuS \cite{WangLLTKW21}. 
While NeuS is not directly applicable in multi-view multi-light acquisition settings in theory, it may become feasible under certain conditions. This feasibility hinges on factors such as the number, spatial distribution and types of lighting conditions, and the object material properties. One can leverage a heuristic method, initially proposed in \cite{LiZWSDT20} and later employed for obtaining pixel depths using MVS in \cite{KayaKOFG22, KayaKOFG23}. This heuristic involves approximating input images captured under mono-illumination for each viewpoint by taking the median of pixel intensities obtained under varying illumination. See, e.g.,~\cite{LiZWSDT20} for detailed information. 

\input{figures/neus_vs_ours_2}
\input{tables/table_CD}

A qualitative comparison between the results of mono-illumination NeuS using this heuristic and the ones from our method is provided in Fig.~\ref{fig:neus_vs_ours}. As can be seen, our proposed approach provides a much finer level of details. In particular, mono-illumination NeuS requires a high number of viewpoints, with a drastic decline in the reconstruction quality when using 5 viewpoints. On the contrary, our method shows stable results, only losing some fine details over concave areas. Moreover, even with all viewpoints used, mono-illumination NeuS fails in reliably reconstructing the low visibility and high curvature areas. 
In addition to Fig. \ref{fig:neus_vs_ours} (right),
this can be observed in the quantitative evaluation provided in Table~\ref{table:cd_all}, where mono-illumination NeuS shows a reconstruction error 62\% higher than ours on low visibility areas and 46\% higher than ours on high curvature areas.

\paragraph{Photometric stereo method.}
Our method can be employed with any PS method.
To illustrate this flexibility, we evaluate the reconstruction accuracy on the Buddha dataset while taking as input the normal maps from CNN-PS~\cite{cnnps18}, used in Kaya22-23~\cite{KayaKOFG22,KayaKOFG23}, and SDPS-Net~\cite{chen2019SDPS_Net}, used in PS-NeRF \cite{YangCCCW22}, in addition to the one obtained using normal maps from SDM-UniPS~\cite{sdmunips} reported in the main paper.
The results are reported in Table \ref{tab:your_label}.
As expected, we observe that the choice of a particular PS technique influences the final outcome, yet our framework consistently improves the results in comparison with previous works, including those based on multi-objective optimizations \cite{KayaKOFG22,KayaKOFG23}.

\begin{table}[h]
\centering
\small
\begin{adjustbox}{width=\linewidth,center}
\begin{tabular}{l|cc||cc||c}
& \multicolumn{2}{c||}{CNN-PS} & \multicolumn{2}{c||}{SDPS-Net} & \multicolumn{1}{c}{SDM-UniPS}\\
Buddha & Kaya23 & Ours & PS-NeRF & Ours & Ours\\ 
\hline
H. curv.  & 0.35 & \textcolor{bestgreen}{\textbf{0.29}} & 0.51 & \textcolor{bestgreen}{\textbf{0.31}} & 0.26\\

Low curv. & 0.24 & \textcolor{bestgreen}{\textbf{0.22}} & 0.33 & \textcolor{bestgreen}{\textbf{0.25}} & 0.23\\
All & 0.25 & \textcolor{bestgreen}{\textbf{0.22}} & 0.34 & \textcolor{bestgreen}{\textbf{0.25}} & 0.23\\ 
\end{tabular}
\end{adjustbox}
\caption{Results of our method with different input normals, namely CNN-PS (used in Kaya22-23), SDPS-Net (used in PS-NeRF) and SDM-UniPS. High curvature corresponds to the results averaged over all the vertices whose absolute curvature is higher than $3.3$.
Our method shows to perform best irrespective of the PS method being used.}
\label{tab:your_label}
\end{table}

\paragraph{Ablation.} We complete our ablation study with qualitative results on the ear and the knot of  Buddha shown in Fig.~\ref{fig:cmp_zoom}.

\input{figures/ablation_buddha}

\paragraph{Additional benchmarking.}
We provide in Fig.~\ref{fig:0} a qualitative comparison of the angular error maps on the five objects of DiLiGenT-MV, for our method and state-of-the-art ones, namely Park16~\cite{ParkSMTK13}, Li19~\cite{LiZWSDT20}, Kaya22~\cite{KayaKSFG22}, PS-NeRF~\cite{YangCCCW22}, Kaya23~\cite{KayaKOFG23}, MVPSNet~\cite{mvpsnet} and also SDM-UniPS~\cite{sdmunips} although it does not provide a full 3D reconstruction.
The recovered geometry shows to be overall more accurate with our method.
Interestingly, our recovered normals overcome the PS ones, especially in concave areas where inter-reflections bias the single-viewpoint reconstruction. Lastly, we provide further quantitative comparisons, namely precision and recall in Fig.~\ref{fig:precision_recall}, and MAE on low visibility and high curvature areas in Table~\ref{table:mae_all}. Our proposed approach consistently yields the most accurate reconstructions.

\input{figures/precision_recall}

\section{Limitations}
\label{sec:limits}
The reconstructions obtained through the proposed method yet exhibit a few poorly reconstructed areas, as illustrated in Fig. \ref{fig:limitation}, particularly for Reading's neck and Bear's right ear.
The suboptimal reconstruction of Reading's neck can be attributed, in part, to inacurracies of the normal estimates from SDM-UniPS.
However, the underlying causes of these discrepancies have yet to be systematically identified.

\input{figures/limitations}

\input{figures/figure_heatmap_MAE}
\input{tables/table_MAE}

%% file: figures/bear_eval_masking.tex
\begin{figure}[!ht]
    \centering
    \begin{tabular}{cc}
        \includegraphics[width=0.4\linewidth]{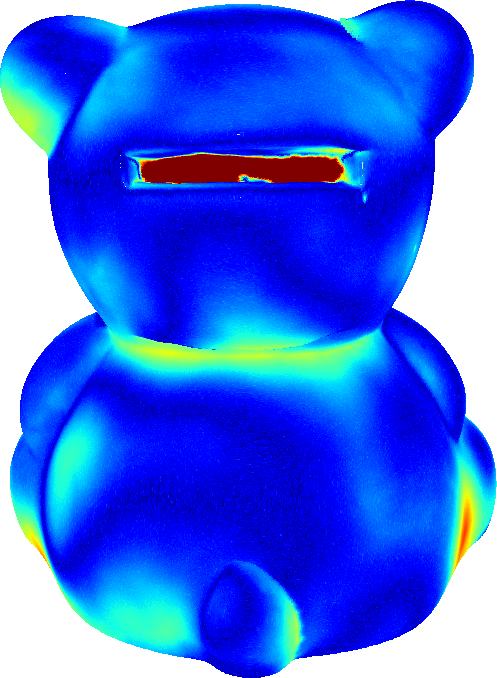} & 
        \includegraphics[width=0.4\linewidth] 
        {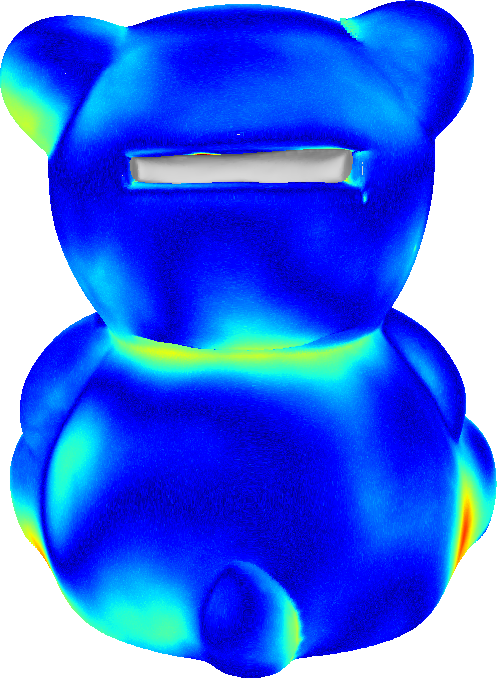} \\
        (a) & (b) \\
    \end{tabular}
    \caption{Rear view of the 3D heatmaps representing errors for the Bear dataset in terms of Chamfer distance.
    (a) The ground truth from DiLiGenT-MV lacks any vertices in the rectangular aperture. For that reason, any method which faithfully reconstructs this area is penalized (area shown in red). This area is thus discarded in all evaluations, providing heatmaps such as (b).}
    \label{fig:bearheat}
\end{figure}

%% file: figures/pot2_section.tex
\begin{figure}[!ht]
    \centering
    \begin{tabular}{cc}
        \hspace{-3mm}
        \includegraphics[width=0.5\linewidth]{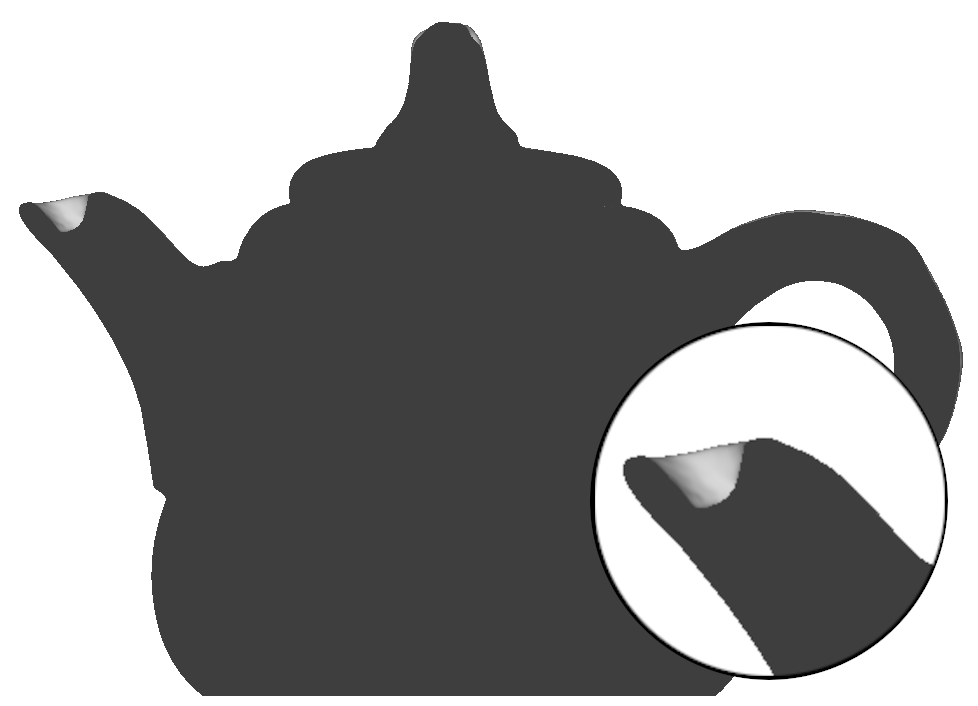} & 
        \hspace{-5mm}\includegraphics[width=0.5\linewidth]{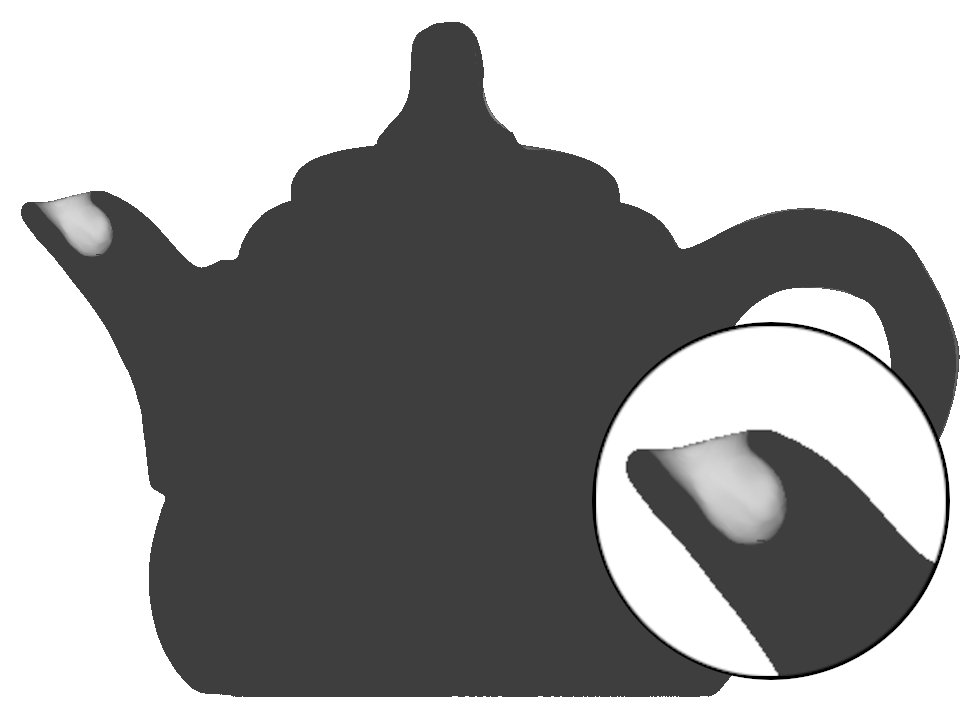} \\
        (a) & (b) \\
    \end{tabular}
    \caption{Cross-section of Pot2's spout delivered by (a) the ground truth of the DiLiGenT-MV dataset and 
    (b) our reconstruction method.
    Our method shows a deeper reconstruction of the internal wall of the spout. This area is thus discarded in all evaluations to avoid penalizing methods that faithfully reconstructs it.}
    \label{fig:pot2section}
\end{figure}

%% file: figures/neus_vs_ours_2.tex
\begin{figure*}[h!]
\centering
\begin{adjustbox}{width=0.99\linewidth,center}
\begin{tabular}{c}
      \def\svgwidth{1.15\linewidth}
      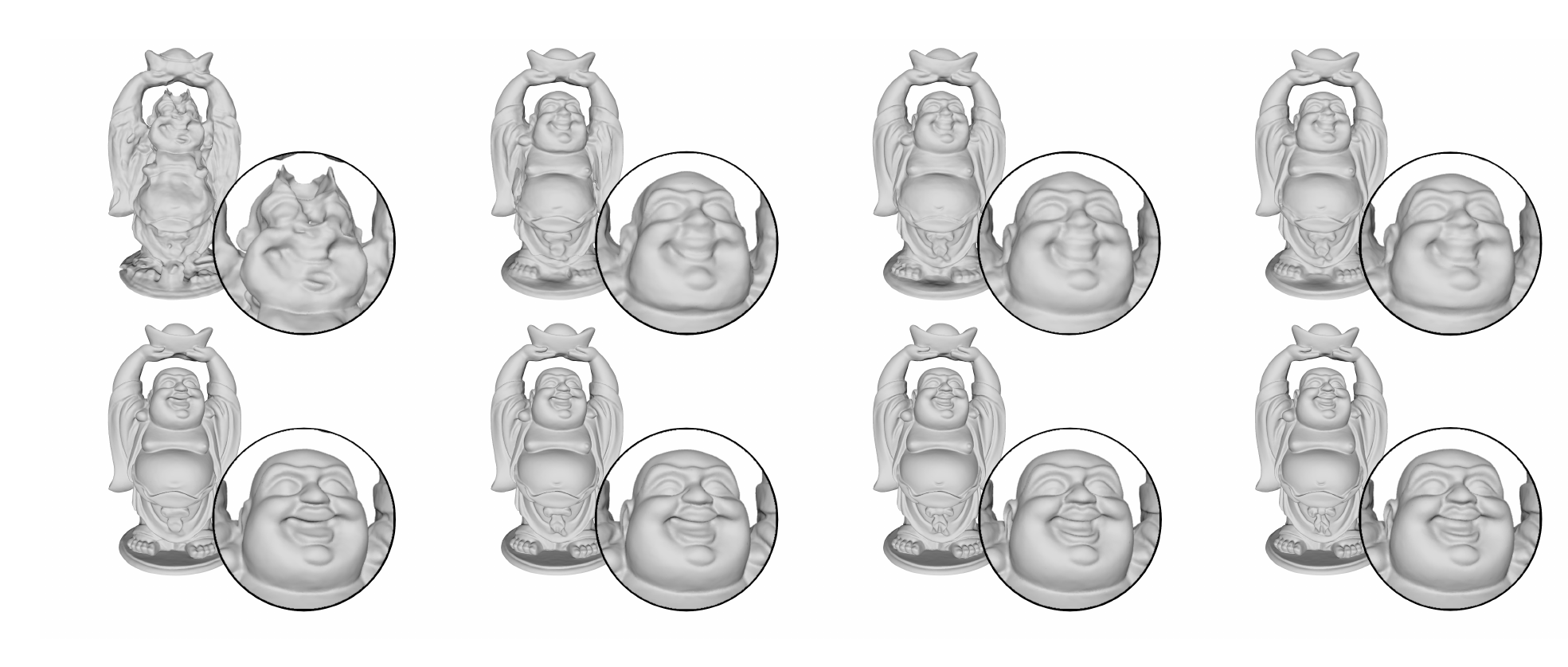  \\
~~~~~~~~~~~------------------------------------------------------------------------------------------------------------------------------------------------~~~~~~~~~~\\
Buddha\\
\vspace{-0.8cm}
\\
      \def\svgwidth{1.15\linewidth}
      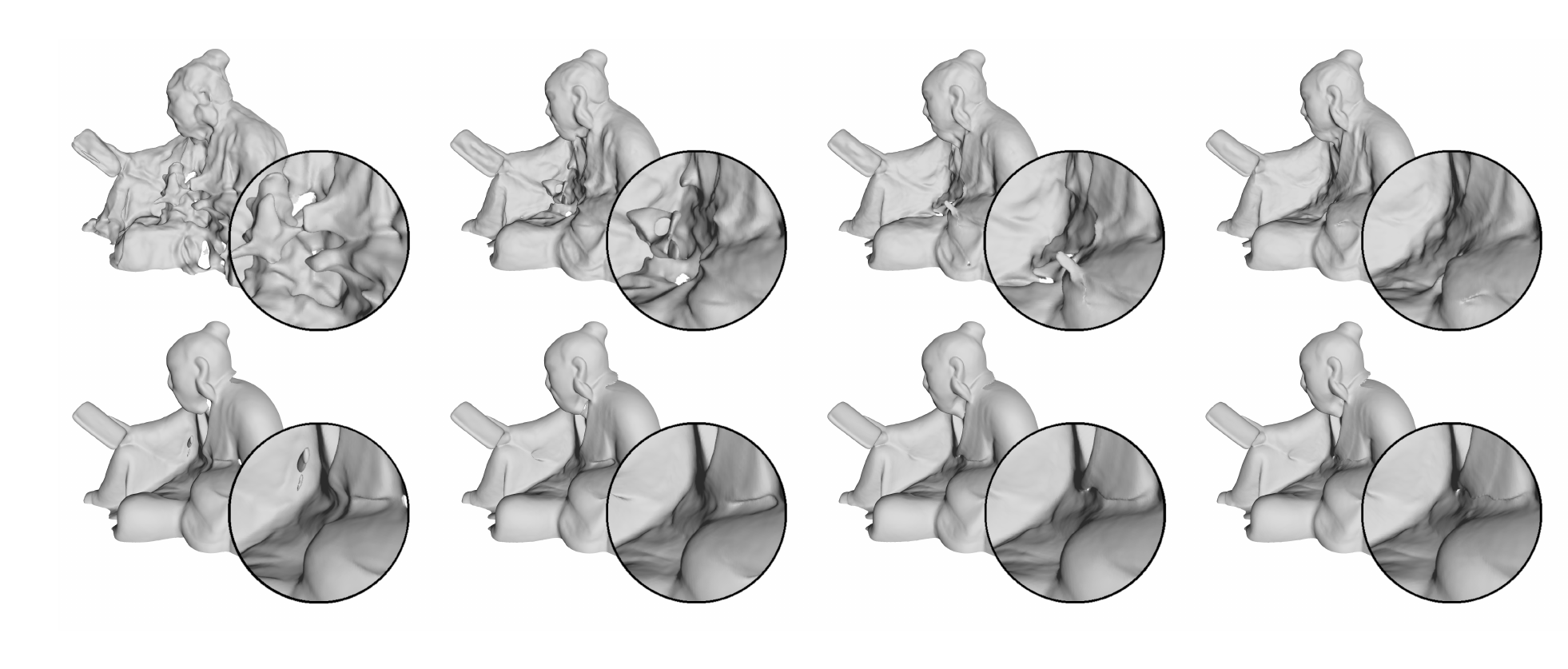  \\
~~~~~~~~~~~------------------------------------------------------------------------------------------------------------------------------------------------~~~~~~~~~~\\
Reading\\
\end{tabular}
\end{adjustbox}
\caption{Qualitative comparison of Buddha and Reading between mono-illumination NeuS and our method, for different number of input viewpoints.}
\label{fig:neus_vs_ours}
\end{figure*}

%% file: figures/buddha_Neus.pdf_tex
\begingroup%
  \makeatletter%
  \providecommand\color[2][]{%
    \errmessage{(Inkscape) Color is used for the text in Inkscape, but the package 'color.sty' is not loaded}%
    \renewcommand\color[2][]{}%
  }%
  \providecommand\transparent[1]{%
    \errmessage{(Inkscape) Transparency is used (non-zero) for the text in Inkscape, but the package 'transparent.sty' is not loaded}%
    \renewcommand\transparent[1]{}%
  }%
  \providecommand\rotatebox[2]{#2}%
  \newcommand*\fsize{\dimexpr\f@size pt\relax}%
  \newcommand*\lineheight[1]{\fontsize{\fsize}{#1\fsize}\selectfont}%
  \ifx\svgwidth\undefined%
    \setlength{\unitlength}{945.08390442bp}%
    \ifx\svgscale\undefined%
      \relax%
    \else%
      \setlength{\unitlength}{\unitlength * \real{\svgscale}}%
    \fi%
  \else%
    \setlength{\unitlength}{\svgwidth}%
  \fi%
  \global\let\svgwidth\undefined%
  \global\let\svgscale\undefined%
  \makeatother%
  \begin{picture}(1,0.42474382)%
    \lineheight{1}%
    \setlength\tabcolsep{0pt}%
    \put(0,0){\includegraphics[width=\unitlength,page=1]{buddha_Neus.pdf}}%
    \put(0.01010682,0.10233322){\rotatebox{90}{\makebox(0,0)[lt]{\lineheight{1.25}\smash{\begin{tabular}[t]{l}$\text{Ours}$\end{tabular}}}}}%
    \put(0.01445811,0.28384226){\rotatebox{90}{\makebox(0,0)[lt]{\lineheight{1.25}\smash{\begin{tabular}[t]{l}$\text{NeuS}$\end{tabular}}}}}%
    \put(0.1007153,0.00216946){\makebox(0,0)[lt]{\lineheight{1.25}\smash{\begin{tabular}[t]{l}$\text{5 views}$\end{tabular}}}}%
    \put(0.33915874,0.00216946){\makebox(0,0)[lt]{\lineheight{1.25}\smash{\begin{tabular}[t]{l}$\text{10 views}$\end{tabular}}}}%
    \put(0.58357036,0.00216946){\makebox(0,0)[lt]{\lineheight{1.25}\smash{\begin{tabular}[t]{l}$\text{15 views}$\end{tabular}}}}%
    \put(0.8294981,0.00216946){\makebox(0,0)[lt]{\lineheight{1.25}\smash{\begin{tabular}[t]{l}$\text{20 views}$\end{tabular}}}}%
  \end{picture}%
\endgroup%

%% file: figures/Reading_Neus.pdf_tex
\begingroup%
  \makeatletter%
  \providecommand\color[2][]{%
    \errmessage{(Inkscape) Color is used for the text in Inkscape, but the package 'color.sty' is not loaded}%
    \renewcommand\color[2][]{}%
  }%
  \providecommand\transparent[1]{%
    \errmessage{(Inkscape) Transparency is used (non-zero) for the text in Inkscape, but the package 'transparent.sty' is not loaded}%
    \renewcommand\transparent[1]{}%
  }%
  \providecommand\rotatebox[2]{#2}%
  \newcommand*\fsize{\dimexpr\f@size pt\relax}%
  \newcommand*\lineheight[1]{\fontsize{\fsize}{#1\fsize}\selectfont}%
  \ifx\svgwidth\undefined%
    \setlength{\unitlength}{941.57494418bp}%
    \ifx\svgscale\undefined%
      \relax%
    \else%
      \setlength{\unitlength}{\unitlength * \real{\svgscale}}%
    \fi%
  \else%
    \setlength{\unitlength}{\svgwidth}%
  \fi%
  \global\let\svgwidth\undefined%
  \global\let\svgscale\undefined%
  \makeatother%
  \begin{picture}(1,0.42632671)%
    \lineheight{1}%
    \setlength\tabcolsep{0pt}%
    \put(0,0){\includegraphics[width=\unitlength,page=1]{Reading_Neus.pdf}}%
    \put(0.01014448,0.10271458){\rotatebox{90}{\makebox(0,0)[lt]{\lineheight{1.25}\smash{\begin{tabular}[t]{l}$\text{Ours}$\end{tabular}}}}}%
    \put(0.01451199,0.28490005){\rotatebox{90}{\makebox(0,0)[lt]{\lineheight{1.25}\smash{\begin{tabular}[t]{l}$\text{NeuS}$\end{tabular}}}}}%
    \put(0.10109063,0.00217754){\makebox(0,0)[lt]{\lineheight{1.25}\smash{\begin{tabular}[t]{l}$\text{5 views}$\end{tabular}}}}%
    \put(0.34042268,0.00217754){\makebox(0,0)[lt]{\lineheight{1.25}\smash{\begin{tabular}[t]{l}$\text{10 views}$\end{tabular}}}}%
    \put(0.58574515,0.00217754){\makebox(0,0)[lt]{\lineheight{1.25}\smash{\begin{tabular}[t]{l}$\text{15 views}$\end{tabular}}}}%
    \put(0.83258938,0.00217754){\makebox(0,0)[lt]{\lineheight{1.25}\smash{\begin{tabular}[t]{l}$\text{20 views}$\end{tabular}}}}%
  \end{picture}%
\endgroup%

%% file: tables/table_CD.tex
\begin{table*}[h!]
\centering
\begin{adjustbox}{width=\linewidth,center}
\begin{tabular}{l|ccccc|c||ccccc|c}
& \multicolumn{6}{c||}{Chamfer distance (visibility 1-5) $\downarrow$} & \multicolumn{6}{c}{Chamfer distance (high curvature) $\downarrow$}\\
Methods & Bear & Buddha & Cow & Pot2 & Reading & Average& Bear & Buddha & Cow & Pot2 & Reading & Average \\
\hline
Park16 &  1.07 & 0.75 & 0.41 & 0.47 & 0.7 & 0.68 & 1.64 & 0.58 & 0.98 & 0.56 & 0.65 & 0.88\\
Li19$\dagger$ &  0.63 & 1.03 & 0.37 & 0.54 & 0.81 & 0.67 & 0.59 & 0.65 & 0.38 & 0.34 & 0.57 & 0.51\\
NeuS &  0.58 & 0.52 & \textcolor{bestgreen}{\textbf{0.17}} & 0.32 & 0.54 & 0.42 & 0.28 & 0.46 & \textcolor{bestblue}{\textbf{0.21}} & 0.39 & 0.38 & 0.35\\
Kaya22 &  0.48 & 0.51 & 0.32 & 0.5 & 0.7 & 0.5 & 0.33 & 0.43 & 0.31 & 0.41 & 0.45 & 0.38\\
PS-NeRF &  0.48 & 0.62 & 0.3 & 0.66 & 0.64 & 0.54 & 0.42 & 0.5 & 0.42 & 0.44 & 0.44 & 0.45\\
Kaya23 &  0.46 & \textcolor{bestblue}{\textbf{0.35}} & 0.39 & 0.42 & \textcolor{bestblue}{\textbf{0.44}} & \textcolor{bestblue}{\textbf{0.41}} & 0.33 & \textcolor{bestblue}{\textbf{0.29}} & \textcolor{bestgreen}{\textbf{0.19}} & \textcolor{bestblue}{\textbf{0.3}} & \textcolor{bestblue}{\textbf{0.33}} & \textcolor{bestblue}{\textbf{0.29}}\\
MVPSNet &  \textcolor{bestblue}{\textbf{0.43}} & 0.68 & 0.27 & 0.49 & 0.57 & 0.49 & 0.56 & 0.58 & 0.52 & 0.47 & 0.54 & 0.53\\
Ours &  \textcolor{bestgreen}{\textbf{0.23}} & \textcolor{bestgreen}{\textbf{0.27}} & \textcolor{bestblue}{\textbf{0.19}} & \textcolor{bestgreen}{\textbf{0.19}} & \textcolor{bestgreen}{\textbf{0.43}} & \textcolor{bestgreen}{\textbf{0.26}} & \textcolor{bestgreen}{\textbf{0.22}} & \textcolor{bestgreen}{\textbf{0.23}} & 0.26 & \textcolor{bestgreen}{\textbf{0.23}} & \textcolor{bestgreen}{\textbf{0.25}} & \textcolor{bestgreen}{\textbf{0.24}}\\
\end{tabular}
\end{adjustbox}
\caption{Chamfer distance on (a) low visibility and (b) high curvature areas. \textcolor{bestgreen}{\textbf{Best results}}. \textcolor{bestblue}{\textbf{Second best results}}.}
\label{table:cd_all}
\end{table*}

%% file: figures/ablation_buddha.tex
\begin{figure}[H]
\centering
\small
\def\svgwidth{\linewidth}
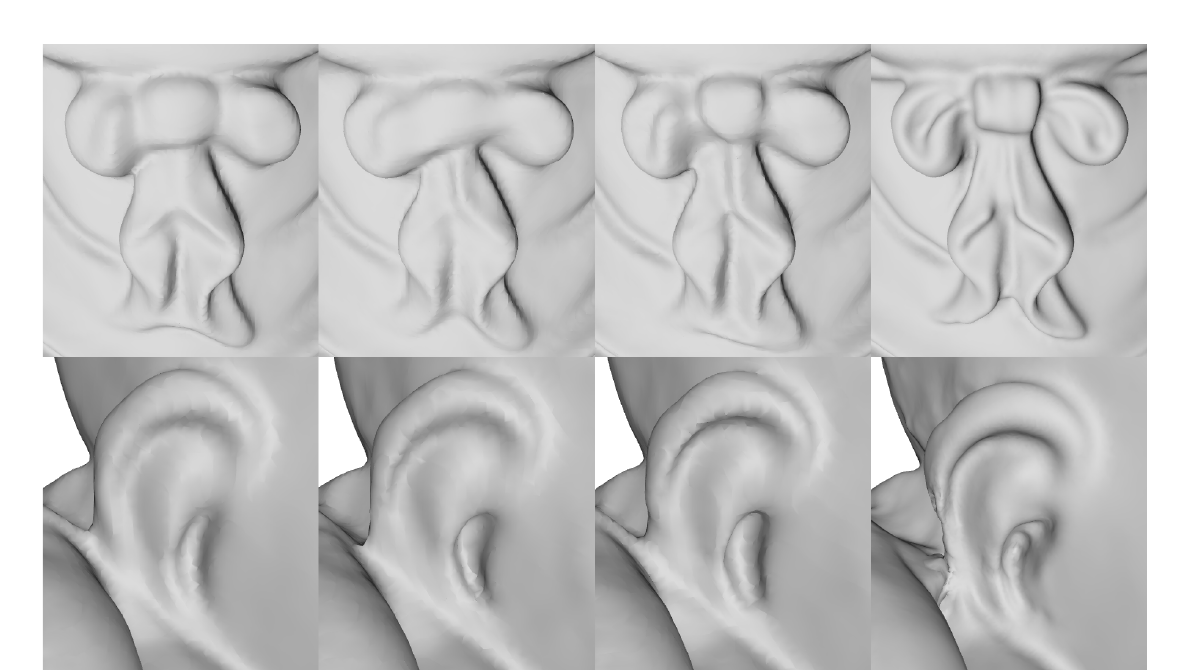

\caption{
Qualitative comparison on the knot and the ear of Buddha between our results and those without the use of reflectance and optimal lighting, disabled individually. Our method exhibits better results in both cases.}
\label{fig:cmp_zoom}
\end{figure}

%% file: figures/ab.pdf_tex
\begingroup%
  \makeatletter%
  \providecommand\color[2][]{%
    \errmessage{(Inkscape) Color is used for the text in Inkscape, but the package 'color.sty' is not loaded}%
    \renewcommand\color[2][]{}%
  }%
  \providecommand\transparent[1]{%
    \errmessage{(Inkscape) Transparency is used (non-zero) for the text in Inkscape, but the package 'transparent.sty' is not loaded}%
    \renewcommand\transparent[1]{}%
  }%
  \providecommand\rotatebox[2]{#2}%
  \newcommand*\fsize{\dimexpr\f@size pt\relax}%
  \newcommand*\lineheight[1]{\fontsize{\fsize}{#1\fsize}\selectfont}%
  \ifx\svgwidth\undefined%
    \setlength{\unitlength}{565.71068482bp}%
    \ifx\svgscale\undefined%
      \relax%
    \else%
      \setlength{\unitlength}{\unitlength * \real{\svgscale}}%
    \fi%
  \else%
    \setlength{\unitlength}{\svgwidth}%
  \fi%
  \global\let\svgwidth\undefined%
  \global\let\svgscale\undefined%
  \makeatother%
  \begin{picture}(1,0.56802855)%
    \lineheight{1}%
    \setlength\tabcolsep{0pt}%
    \put(0,0){\includegraphics[width=\unitlength,page=1]{ab.pdf}}%
    \put(0.05921278,0.54574959){\makebox(0,0)[lt]{\lineheight{1.25}\smash{\begin{tabular}[t]{l}$\text{W/o reflect.}$\end{tabular}}}}%
    \put(0.27520231,0.54638931){\makebox(0,0)[lt]{\lineheight{1.25}\smash{\begin{tabular}[t]{l}$\text{W/o opt. light.}$\end{tabular}}}}%
    \put(0.58146853,0.54574959){\makebox(0,0)[lt]{\lineheight{1.25}\smash{\begin{tabular}[t]{l}$\text{Ours}$\end{tabular}}}}%
    \put(0.82562058,0.54574959){\makebox(0,0)[lt]{\lineheight{1.25}\smash{\begin{tabular}[t]{l}$\text{GT}$\end{tabular}}}}%
    \put(0.02163924,0.36062423){\rotatebox{90}{\makebox(0,0)[lt]{\lineheight{1.25}\smash{\begin{tabular}[t]{l}$\text{Knot}$\end{tabular}}}}}%
    \put(0.02163924,0.10345841){\rotatebox{90}{\makebox(0,0)[lt]{\lineheight{1.25}\smash{\begin{tabular}[t]{l}$\text{Ear}$\end{tabular}}}}}%
  \end{picture}%
\endgroup%

%% file: figures/precision_recall.tex
\begin{figure}[H]
\centering
\begin{tabular}{cc}
\hspace{-3mm}\includegraphics[width = 0.4995\linewidth]{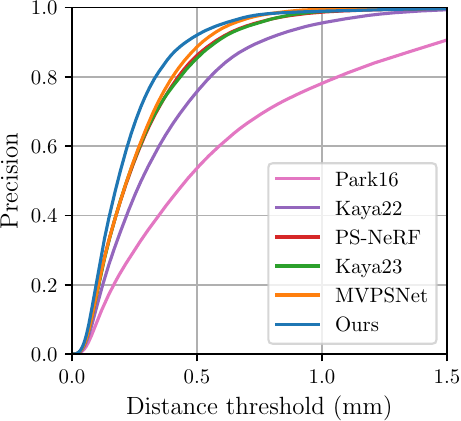} &
\hspace{-2.5mm}\includegraphics[width = 0.5005\linewidth]{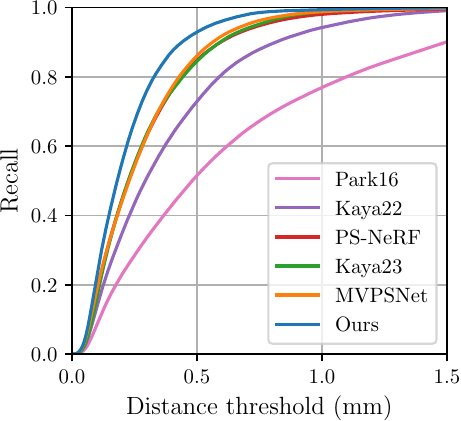} \\
{\small (a)} & {\small (b)} \vspace*{-.5em}
\end{tabular}
\caption{(a) Precision (higher is better) and (b) recall (higher is better) as functions
of the distance error threshold, in comparison with other state-of-the-art methods. 
}
\label{fig:precision_recall}
\end{figure}

%% file: figures/limitations.tex
\begin{figure}[]
    \centering
    \begin{tabular}{cc}
        \hspace{-3mm}
        \includegraphics[height=0.5\linewidth]{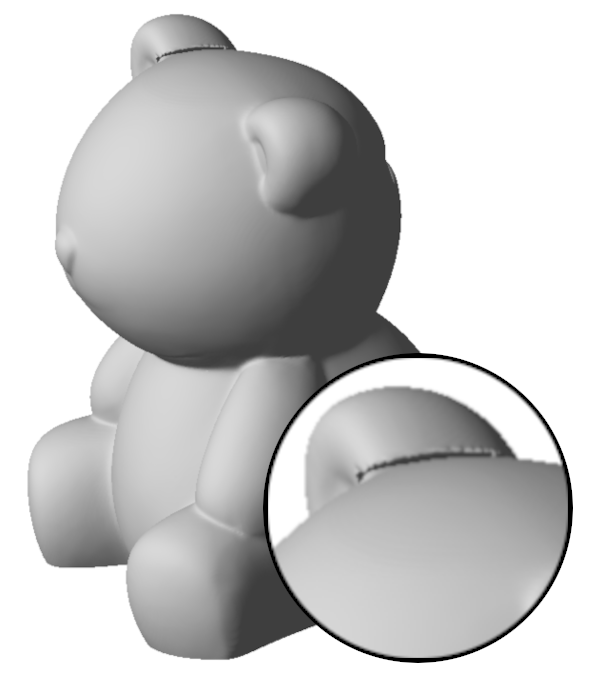} & 
        \includegraphics[height=0.5\linewidth]{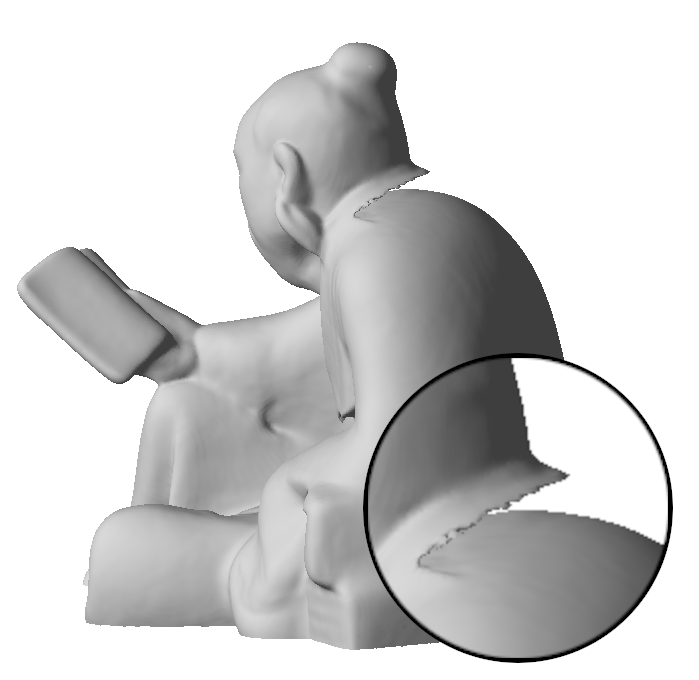} \\
        (a) & (b) \\
    \end{tabular}
    \caption{Regions in Bear (a) and Reading (b) where our method exhibits limitations.}
    \label{fig:limitation}
\end{figure}

%% file: figures/figure_heatmap_MAE.tex
\begin{figure*}[!ht]
\centering
\begin{adjustbox}{width=1.0\linewidth,center}
\begin{tabular}{cccccccc}
\includegraphics[width = 0.1\linewidth]{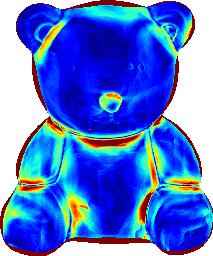} &
\includegraphics[width = 0.1\linewidth]{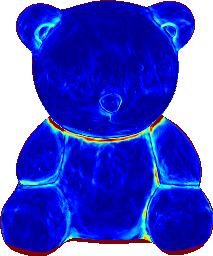} &
\includegraphics[width = 0.1\linewidth]{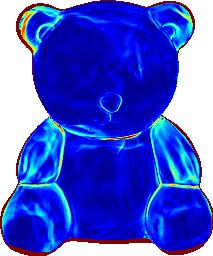} &
\includegraphics[width = 0.1\linewidth]{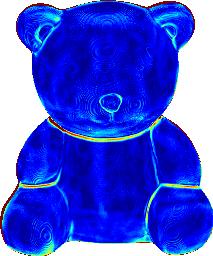} &
\includegraphics[width = 0.1\linewidth]{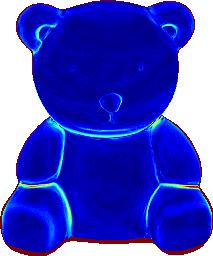} &
\includegraphics[width = 0.1\linewidth]{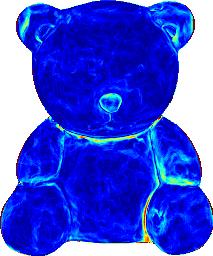} &
\includegraphics[width = 0.1\linewidth]{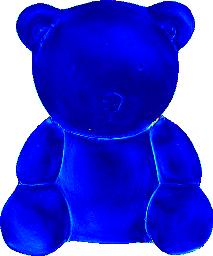} &
\includegraphics[width = 0.1\linewidth]{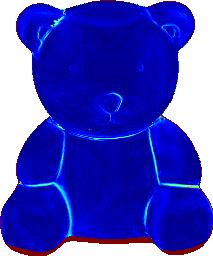}\\
\includegraphics[width = 0.1\linewidth]{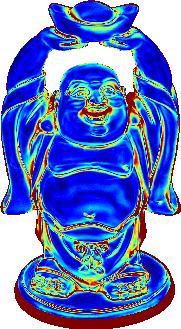} &
\includegraphics[width = 0.1\linewidth]{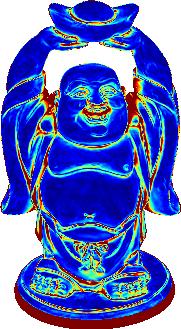} &
\includegraphics[width = 0.1\linewidth]{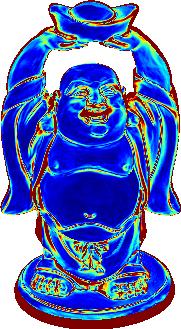} &
\includegraphics[width = 0.1\linewidth]{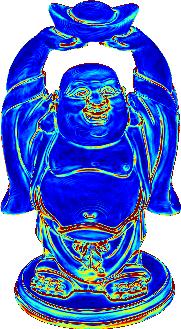} &
\includegraphics[width = 0.1\linewidth]{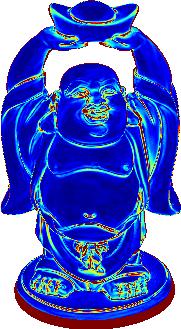} &
\includegraphics[width = 0.1\linewidth]{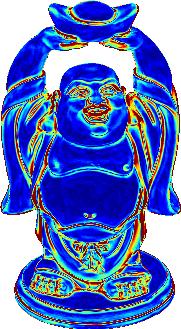} &
\includegraphics[width = 0.1\linewidth]{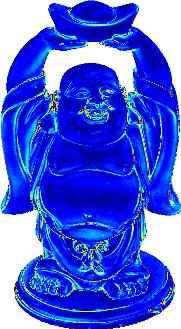} &
\includegraphics[width = 0.1\linewidth]{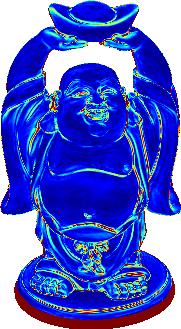}\\
\includegraphics[width = 0.1\linewidth]{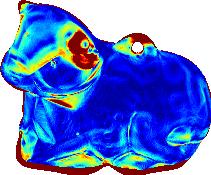} &
\includegraphics[width = 0.1\linewidth]{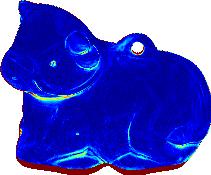} &
\includegraphics[width = 0.1\linewidth]{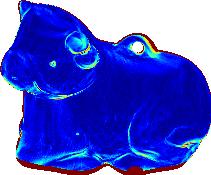} &
\includegraphics[width = 0.1\linewidth]{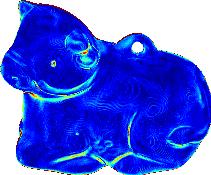} &
\includegraphics[width = 0.1\linewidth]{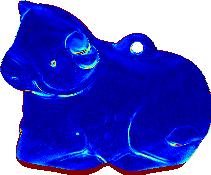} &
\includegraphics[width = 0.1\linewidth]{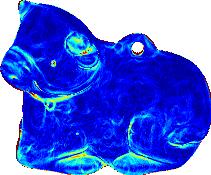} &
\includegraphics[width = 0.1\linewidth]{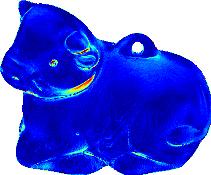} &
\includegraphics[width = 0.1\linewidth]{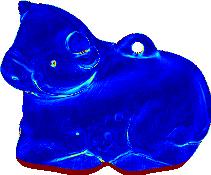}\\
\includegraphics[width = 0.1\linewidth]{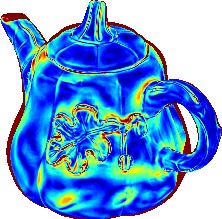} &
\includegraphics[width = 0.1\linewidth]{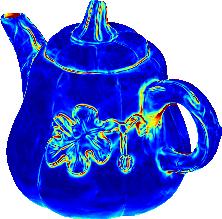} &
\includegraphics[width = 0.1\linewidth]{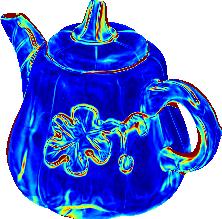} &
\includegraphics[width = 0.1\linewidth]{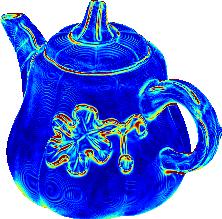} &
\includegraphics[width = 0.1\linewidth]{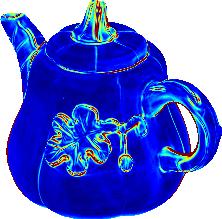} &
\includegraphics[width = 0.1\linewidth]{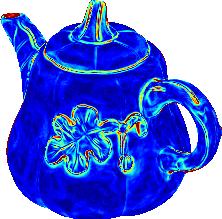} &
\includegraphics[width = 0.1\linewidth]{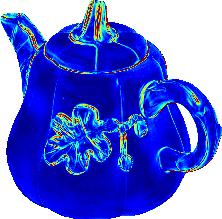} &
\includegraphics[width = 0.1\linewidth]{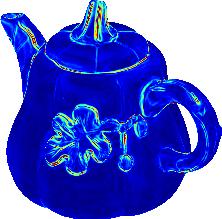}\\
\includegraphics[width = 0.1\linewidth]{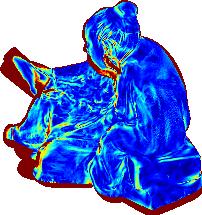} &
\includegraphics[width = 0.1\linewidth]{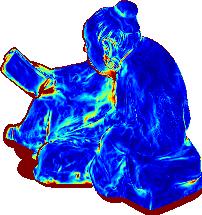} &
\includegraphics[width = 0.1\linewidth]{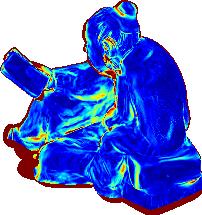} &
\includegraphics[width = 0.1\linewidth]{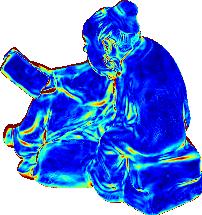} &
\includegraphics[width = 0.1\linewidth]{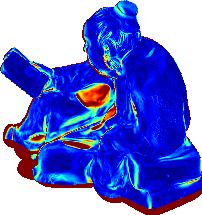} &
\includegraphics[width = 0.1\linewidth]{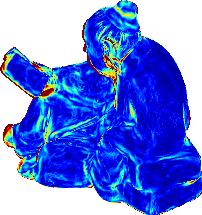} &
\includegraphics[width = 0.1\linewidth]{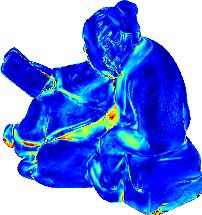} &
\includegraphics[width = 0.1\linewidth]{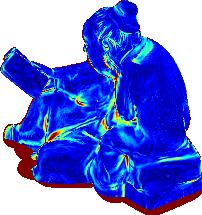}\\
Park16 & Li19$\dagger$ & Kaya22 & PS-NeRF & Kaya23 & MVPSNet & SDM & Ours
\end{tabular}
\end{adjustbox}
\caption{Normal angular error comparison over all DiLiGenT-MV dataset between state-of-the-art methods and ours.}
\label{fig:0}
\end{figure*}

%% file: tables/table_MAE.tex
\begin{table*}[h!]
\centering
\begin{adjustbox}{width=\linewidth,center}
\begin{tabular}{l|ccccc|c||ccccc|c}
& \multicolumn{6}{c||}{Normal MAE (Visibility 1-5) $\downarrow$} & \multicolumn{6}{c}{Normal MAE (High Curvature) $\downarrow$}\\
Methods & Bear & Buddha & Cow & Pot2 & Reading & Average& Bear & Buddha & Cow & Pot2 & Reading & Average \\
\hline
Park16 &  38.5 & 29.3 & 34.6 & \textcolor{bestblue}{\textbf{25.2}} & 20.6 & 29.6 & 31.7 & 26.2 & 39.5 & \textcolor{bestblue}{\textbf{23.3}} & 24.1 & 29.0\\
Li19$\dagger$ &  41.1 & 33.7 & 29.4 & 39.0 & 23.3 & 33.3 & 26.5 & 26.4 & 30.6 & 23.4 & 24.1 & 26.2\\
Kaya22 &  32.0 & 27.6 & 40.5 & 40.0 & 18.4 & 31.7 & 20.2 & 29.1 & 35.9 & 32.8 & 21.8 & 28.0\\
PS-NeRF &  19.4 & 19.6 & 27.4 & 32.2 & 21.1 & 24.0 & 21.2 & 28.0 & 27.9 & 23.9 & 28.3 & 25.8\\
Kaya23 &  19.4 & 17.6 & \textcolor{bestgreen}{\textbf{24.0}} & 28.1 & \textcolor{bestblue}{\textbf{14.6}} & 20.7 & 24.1 & 24.2 & \textcolor{bestgreen}{\textbf{21.6}} & 28.5 & \textcolor{bestgreen}{\textbf{19.3}} & \textcolor{bestblue}{\textbf{23.6}}\\
MVPSNet &  31.1 & 29.6 & 30.4 & 35.3 & 18.1 & 28.9 & \textcolor{bestblue}{\textbf{18.7}} & 26.3 & 27.7 & \textcolor{bestgreen}{\textbf{23.3}} & 23.5 & 23.9\\
SDM &  \textcolor{bestgreen}{\textbf{12.9}} & \textcolor{bestblue}{\textbf{14.4}} & 28.5 & 25.7 & 16.9 & \textcolor{bestblue}{\textbf{19.7}} & 21.6 & \textcolor{bestgreen}{\textbf{21.0}} & \textcolor{bestblue}{\textbf{23.4}} & 28.2 & 24.7 & 23.8\\
Ours &  \textcolor{bestblue}{\textbf{13.0}} & \textcolor{bestgreen}{\textbf{14.1}} & \textcolor{bestblue}{\textbf{26.8}} & \textcolor{bestgreen}{\textbf{21.5}} & \textcolor{bestgreen}{\textbf{13.5}} & \textcolor{bestgreen}{\textbf{17.8}} & \textcolor{bestgreen}{\textbf{18.4}} & \textcolor{bestblue}{\textbf{24.2}} & 28.0 & 24.9 & \textcolor{bestblue}{\textbf{19.9}} & \textcolor{bestgreen}{\textbf{23.1}}\\
\end{tabular}
\end{adjustbox}
\caption{Normal MAE on (a) low visibility and (b) high curvature areas. \textcolor{bestgreen}{\textbf{Best results}}. \textcolor{bestblue}{\textbf{Second best results}}.}
\label{table:mae_all}
\end{table*}